\documentclass{article}

\usepackage[preprint]{neurips_2026}

\usepackage[utf8]{inputenc} 
\usepackage[T1]{fontenc}    
\usepackage{hyperref}       
\usepackage{url}            
\usepackage{booktabs}       
\usepackage{amsfonts}       
\usepackage{nicefrac}       
\usepackage{microtype}      
\usepackage{xcolor}         

\usepackage{amssymb}
\usepackage{booktabs}
\usepackage{multirow}
\usepackage{bm}
\usepackage{bbding}
\usepackage{algorithm}
\usepackage{algorithmic}
\usepackage{amsmath}

\usepackage{tcolorbox}
\usepackage{colortbl}
\usepackage{multirow}
\usepackage{graphicx}
\usepackage{wrapfig}
\usepackage{caption}
\usepackage{subcaption}
\usepackage{enumitem}

\definecolor{citecolor}{HTML}{0071bc}

\usepackage{authblk}

\usepackage{soul}

\newcommand{\ourMethod}{StreamPI}

\title{\ourMethod: Streaming Multimodal Temporal Modeling for Vision-Language-Action Models}

\author[ ]{\textbf{Zhe Liu}$^{1*}$}
\author[ ]{\textbf{Jinghua Hou}$^{1}$\thanks{Equal contribution.}}
\author[ ]{\textbf{Yuxiang Lu}$^{1}$}
\author[ ]{\textbf{Zhenya Yang}$^{1}$}
\author[ ]{\textbf{Xianzhe Fan}$^{1}$}
\author[ ]{\textbf{Junwei Luo}$^{1}$}
\author[ ]{\textbf{Junyi Li}$^{1}$}
\author[ ]{\textbf{Ruihua Han}$^{1}$}
\author[ ]{\textbf{Zhi Hou}$^{2}$}
\author[ ]{\textbf{Hengshuang Zhao}$^{1}$\thanks{Corresponding author.}}
\affil[1]{The University of Hong Kong} 
\affil[2]{ACE Robotics \ \ }
\affil[ ]{\textcolor{magenta}
{\url{https://happinesslz.github.io/projects/StreamPI}}}

\begin{document}

\maketitle

\begin{abstract}

Vision-Language-Action (VLA) models have demonstrated effectiveness in robot manipulation, yet state-of-the-art models such as $\pi_{0.5}$ operate under a single-frame paradigm, limiting their ability to retain past observations and develop precise spatial perception.
In this paper, we propose \ourMethod, a streaming multimodal temporal modeling framework that equips single-frame VLA with temporal reasoning capability without introducing any additional parameters.
One core design is \textit{instruction-anchored temporal modeling}. It treats each (visual observation, language instruction) pair as an atomic temporal unit: bidirectional attention within each pair enables cross-modal fusion, while causal attention across pairs preserves autoregressive streaming inference.
This ensures the language instruction serves as a persistent semantic anchor throughout task execution.
To bridge the gap between synchronous training and asynchronous real-robot deployment, we introduce a \textit{random-interval streaming training} strategy: a proper inter-frame interval~(e.g., every 3 frames) enables faster and smoother action execution. Beyond this, randomizing the interval further improves robustness to frame-timing perturbations, supporting asynchronous deployment in practice.
Furthermore, by leveraging the length extrapolation capability of the LLM backbone, \ourMethod~seamlessly inherits pretrained single-frame weights and supports flexible single-frame and multi-frame inference.
Experiments on real-robot tasks spanning  memory-dependent and precise perception scenarios, as well as the simulation benchmark LIBERO,
demonstrate that {\ourMethod} outperforms $\pi_{0.5}$ across diverse tasks.

\end{abstract}

\section{Introduction}

\begin{figure*}[h!]
\centering
\includegraphics[width=0.95\linewidth]{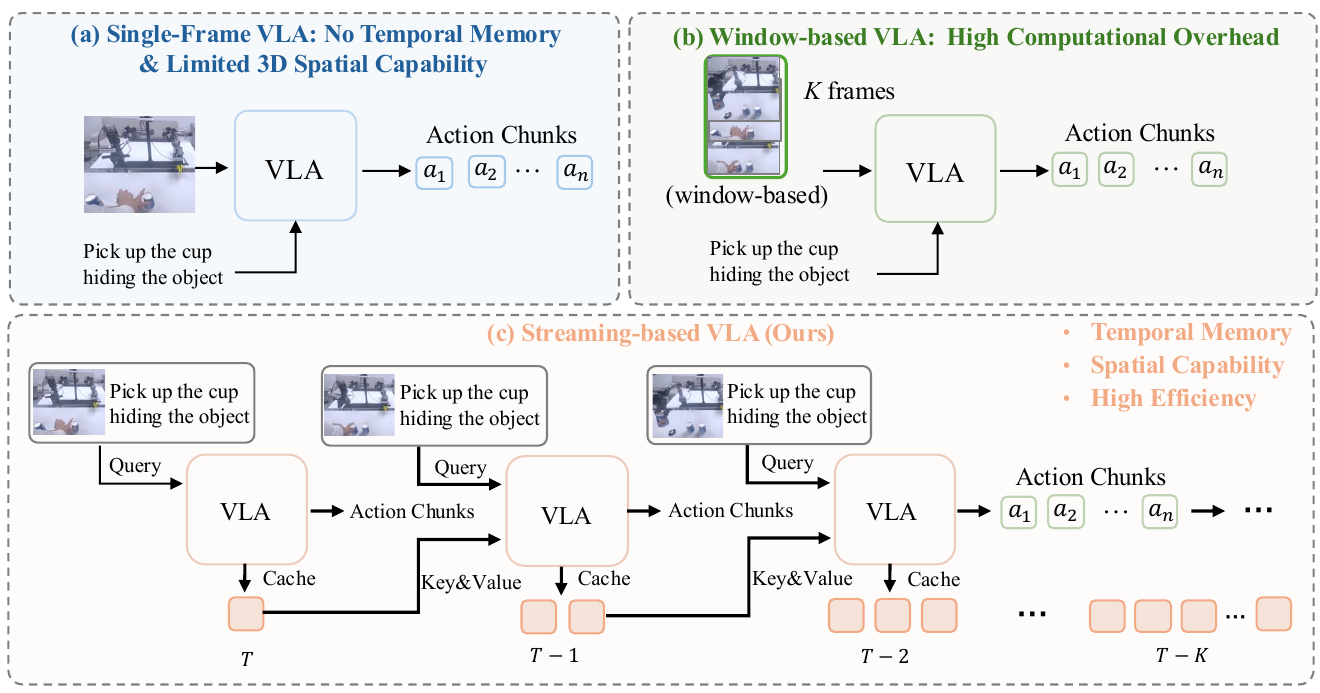}
\vspace{-5pt}
\caption{
Comparison of three paradigms for VLA-based robot manipulation.
\textbf{(a) Single-frame VLA:} Only the current observation is processed, lacking historical context and limiting temporal memory and spatial perception.
\textbf{(b) Window-based VLA:} A window of $K$ frames is processed simultaneously, enriching temporal context at the cost of high computational overhead.
\textbf{(c) Streaming-based VLA (Ours):} The model queries cached Key\&Value representations from previous timesteps via a lightweight KV cache, achieving temporal memory and precise spatial perception. 
}
\label{fig:intro}
\vspace{-15pt}
\end{figure*}

Vision-Language-Action (VLA) models have emerged as a promising 
paradigm for generalizable robot manipulation, unifying perception, 
language understanding, and action generation within a single 
end-to-end framework~\cite{rt2, openvla, pi0, pi05}. 
Despite their impressive performance, state-of-the-art VLA models
such as $\pi_0$ and $\pi_{0.5}$ operate under a single-frame paradigm,
as illustrated in Figure~\ref{fig:intro}(a), where each action is
predicted from a single image observation without access to any
historical context.
This design inherently precludes two critical capabilities,
namely the ability to memorize and reason over past observations,
and the capacity to develop precise spatial perception that
emerges from temporal aggregation.

Incorporating temporal context addresses both limitations 
simultaneously. On the memory side, access to historical observations 
enables robots to resolve tasks that are inherently ambiguous from a 
single frame, such as inferring which cup conceals a target object or 
grasping a dynamically moving target. On the perception side, the 
benefit of temporal modeling for spatial understanding has been well 
established in autonomous driving, where multi-frame fusion 
methods~\cite{bevdet4d, bevformer, wang2023exploring,open} dramatically improve 
3D scene understanding, and VGGT~\cite{vggt,zhuo2025streaming} series demonstrate that 
temporal aggregation yields substantially more accurate depth 
estimation and 3D reconstruction. These findings transfer naturally 
to embodied manipulation, where strong spatial perception is 
fundamental to precise object grasping and placement. 
To this goal, shown in Figure~\ref{fig:intro}(b), some multiple-frame VLA methods~\cite{dynamicvla,torne2026mem} adopt window-based visual inputs
for temporal modeling.
To reduce inference time, such approaches
typically resort to either a smaller backbone network
or a dedicated video encoder that compresses multi-frame
observations into a reduced set of tokens.

Therefore, effectively incorporating temporal 
information into strong single-frame VLA foundation models such as 
$\pi_{0.5}$ is non-trivial. There are four potential challenges.
\textbf{1) Computational overhead:}
Concatenating frames across time causes sequence length to grow 
linearly, rendering inference latency prohibitive 
for real-time control. 
\textbf{2) Instruction forgetting:}
Conventional VLA models inject the language instruction as a fixed set of text tokens. As the temporal horizon expands, accumulating visual tokens 
progressively dilute the influence of instruction tokens, causing the 
model to lose track of the task goal over long horizons.
\textbf{3) Training-deployment mismatch:} 
Training relies on regularly sampled frame sequences, whereas real-robot deployment produces asynchronous observation streams with variable time gaps,
degrading model robustness in online settings.
\textbf{4) Representation corruption:}
Introducing a video encoder adds new 
parameters whose feature distributions are misaligned with the powerful VLA
pretrained models, risking corruption of the 
rich visual-language features of the base model in the embodied domain.

To address these challenges, we propose \ourMethod, a streaming
multi-modal temporal modeling framework,
as illustrated in Figure~\ref{fig:intro}(c).
\ourMethod~incorporates four key designs to enable efficient
and robust temporal reasoning.
\textbf{1) Streaming inference:} Rather than processing a full
observation window at every step, \ourMethod~adopts a streaming
temporal modeling paradigm that caches Key\&Value representations
from past timesteps, keeping inference cost constant regardless
of temporal horizon.
\textbf{2) Instruction-anchored temporal modeling:} Rather than
modeling temporal dependencies over visual frames alone, \ourMethod~treats
each (visual observation, language instruction) pair as an atomic
temporal unit. The language instruction is persistently coupled
with every observation as a semantic anchor throughout execution.
Within each pair, bidirectional attention enables thorough
cross-modal fusion, while causal attention across pairs preserves
the autoregressive structure for online streaming inference,
ensuring the model consistently maintains awareness of the
current task goal.
\textbf{3) Random-interval streaming training:} To improve
robustness to variable frame rates and asynchronous observation
arrival at deployment time, \ourMethod~adopts a random-interval
streaming training strategy that exposes the model to diverse
frame-timing perturbations during training.
\textbf{4) No additional parameters:} By leveraging the
length extrapolation capability of the LLM backbone, \ourMethod~
seamlessly inherits all pretrained weights of $\pi_{0.5}$ without
introducing any new parameters, fully preserving its
representational integrity while naturally supporting both
single-frame and multi-frame inference at test time.

Finally, we evaluate {\ourMethod} on real-robot tasks spanning 
\textit{precise perception-dependent tasks} and \textit{memory-dependent tasks} and consistently outperforms $\pi_{0.5}$. On the LIBERO simulation benchmark~\cite{libero}, 
{\ourMethod} achieves superior performance, verifying the effectiveness of our approach.

In summary, our main contributions are as follows:
\begin{itemize}[leftmargin=*]
    \item We propose \ourMethod, a streaming multi-modal temporal modeling VLA framework that treats each
    (visual observation, language instruction) pair as an atomic temporal unit with instruction-anchored modeling to maintain persistent task awareness during execution.

    \item We introduce a random-interval streaming training strategy
    that exposes the model to diverse frame-timing perturbations,
    bridging the gap between synchronous training and asynchronous
    real-robot deployment. Besides, \ourMethod~seamlessly inherits
    all pretrained weights of $\pi_{0.5}$ via LLM length extrapolation and supports flexible single-frame and multi-frame inference.

    \item Extensive experiments on real-robot manipulation and
    the LIBERO benchmark demonstrate that \ourMethod~consistently
    outperforms $\pi_{0.5}$ on both spatial-precision and
    memory-dependent tasks.
\end{itemize}
\section{Related Work}

\noindent \textbf{Vision-Language-Action Models.}
The emergence of large-scale pre-trained vision-language models (VLMs) has catalyzed a new generation of robot policies that unify perception, language understanding, and action generation within a single framework~\cite{rt2, openvla, pi0, gong2026ace,team2026ace,pi05,ni2025swiftvla,xvla,univla,vlaadapter,lingbotvla,tinyvla,chen2025nanovla,spatialvla,liu2026drivepi,cen2025rynnvla,zhang2025dreamvla}.
RT-2~\cite{rt2} pioneered this direction by co-finetuning a VLM on robot demonstration data, demonstrating that web-scale visual-linguistic knowledge can be directly transferred to robotic control.
OpenVLA~\cite{openvla} extended this paradigm with an open-source framework, enabling broader community adoption and systematic study of VLA design choices.
More recently, $\pi_0$~\cite{pi0} introduced a flow-matching action head decoupled from the VLM backbone, achieving high-frequency dexterous control while preserving the semantic reasoning capabilities of the language model.
$\pi_{0.5}$~\cite{pi05} further scales this approach with improved data diversity and task generalization.
Despite these advances, existing VLA models predominantly operate in a {single-frame} inference paradigm, processing each observation independently without maintaining temporal context across time steps.
This motivates our work on streaming temporal modeling for VLA inference.

\noindent \textbf{Temporal Modeling for Robot Manipulation.}
Incorporating temporal context into robot policies has long been recognized as essential for tasks requiring spatial reasoning and long-horizon planning~\cite{chi2025diffusion,act}.
Diffusion Policy~\cite{chi2025diffusion} and ACT~\cite{act} demonstrate that conditioning on short observation histories significantly improves action consistency and task success rates.
Recent work~\cite{dynamicvla,li2026towards,shi2025memoryvla,ma2026st,zhang20254d,li2026remem,jang2025contextvla,zheng2024tracevla} has explored multi-frame modeling within VLA frameworks. CronusVLA~\cite{li2026towards} systematically investigates the design space of multi-frame VLA models, showing that incorporating historical observations yields substantial gains on manipulation benchmarks requiring spatial precision.
World model approaches~\cite{fung2025embodied, li2025comprehensive,gao2026dreamdojo,song2026fast,yuan2026fast,bi2025motus} further argue that temporal modeling is indispensable for agents to anticipate future states and reason about action consequences.
However, these methods either introduce significant computational overhead through full attention over all historical tokens, or decouple visual observations from their corresponding language instructions during temporal aggregation, which is a design flaw that leads to instruction forgetting over long task horizons.
{\ourMethod} addresses both limitations through its instruction-anchored atomic temporal unit design, which preserves cross-modal binding while enabling efficient causal streaming inference.

\noindent \textbf{Streaming Inference.}
The challenge of efficient streaming inference has been extensively studied in the NLP community.
StreamingLLM~\cite{xiao2023efficient} identifies the ``attention sink'' phenomenon and proposes retaining a small set of initial tokens alongside a sliding window of recent tokens, enabling LLMs to process infinitely long sequences without recomputation.
LongLoRA~\cite{chen2023longlora}  demonstrates that transformer models can generalize to sequence lengths far beyond those seen during training. 
In the robotics domain, recent work~\cite{duan2025real,wang2026real,lu2026faster,shi2026streamingvla} on asynchronous execution highlights the practical challenge of variable-rate observation streams, where fixed-interval assumptions break down under real-world deployment conditions.
Our random-interval streaming training strategy directly addresses this gap, exposing the model to diverse temporal spacings during training to improve robustness to the asynchronous observation streams encountered on physical robots.

\section{Method}

\begin{figure*}[t!]
\centering
\includegraphics[width=0.99\linewidth]{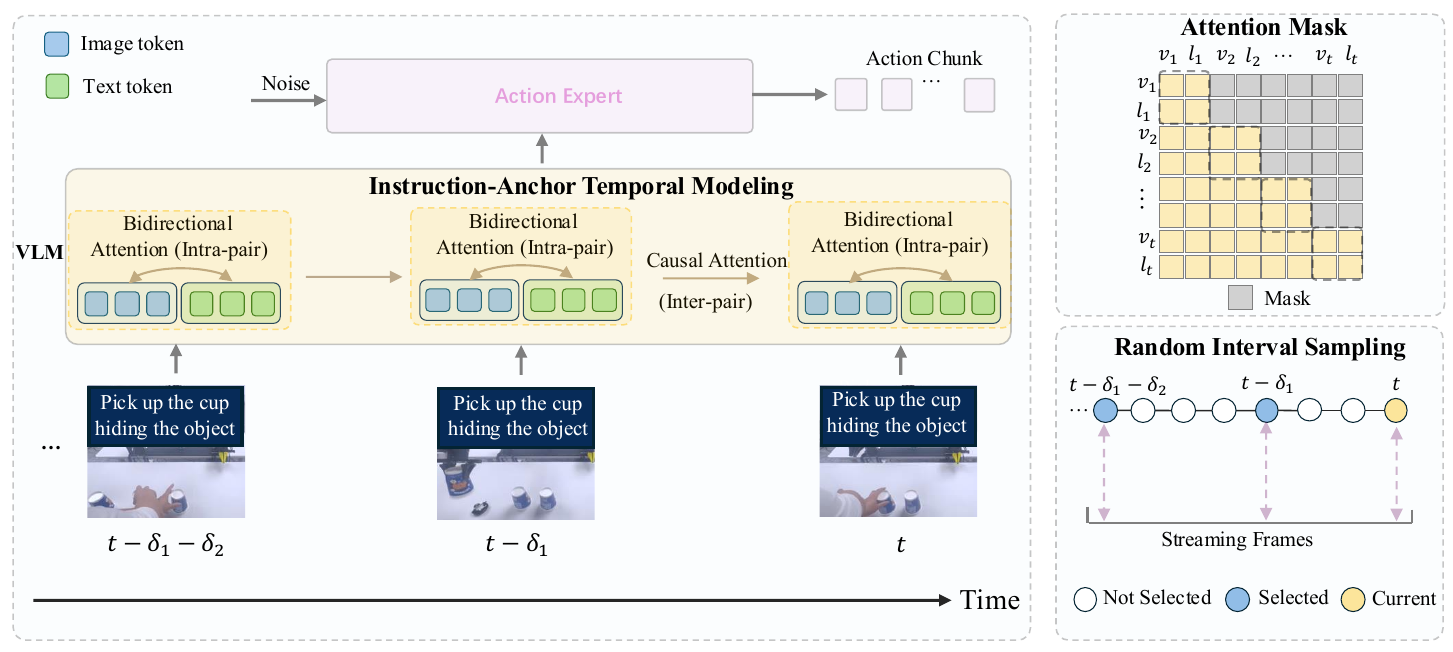}
\vspace{-5pt}
\caption{The pipeline of {\ourMethod}. To fully unleash the potential of multi-modal interaction, we use bidirectional attention for image-text pairs and causal attention for inter-frames with block-wise causal attention mask. We additional use the random interval sampling to improve the temporal robustness in the real-world deployment.}
\label{fig:method}
\vspace{-10pt}
\end{figure*}

In this section, we present {\ourMethod}, a streaming multi-modal temporal modeling framework for Vision-Language-Action models~(\textit{e.g.}, $\pi_{0.5}$), as illustrated in Figure~\ref{fig:method}.
We begin by revisiting the single-frame inference paradigm of $\pi_{0.5}$ and identifying its key limitations (Sec.~\ref{sec:preliminaries}).
We then introduce our core architectural design, which encapsulates visual observations and language instructions into \textit{(image, text)} pairs as atomic temporal units, and organizes them through a causal attention mechanism (Sec.~\ref{sec:temporal_modeling}).
Finally, we describe our random-interval streaming training strategy, which bridges the gap between training and asynchronous real-robot deployment (Sec.~\ref{sec:training}).

\subsection{Preliminaries}
\label{sec:preliminaries}

$\pi_{0.5}$ is a Vision-Language-Action (VLA) model that processes a
single observation at each inference step.
At time $t$, the model receives a language instruction ${l}$
and a set of multi-view visual observations $\mathbf{V}_t =
\{{v}_t^f,\, {v}_t^l,\, {v}_t^r\}$, comprising
a front-view, left-view and right-view wrist camera images,
where each of  ${v_t^f, v_t^l, v_t^r} \in \mathbb{R}^{H \times W \times 3}$.
These inputs are tokenized and concatenated into a sequence: $\mathbf{x}_t = [\mathbf{V}_t, {l}_t]$.
The model then samples an action $\mathbf{a}_t$ from the learned
policy conditioned on $\mathbf{x}_t$: $\mathbf{a}_t \sim \pi_\theta(\cdot \mid \mathbf{x}_t)$, where $\pi_\theta$ denotes the policy parameterized by the
pre-trained weights.
$\pi_{0.5}$ employs a transformer backbone with full
bidirectional attention over the input tokens, enabling rich
cross-modal fusion between the language instruction and all
visual observations within a single frame.

However, a single-frame observation provides insufficient geometric context for tasks requiring precise spatial perception, whereas temporal observations can reveal richer latent geometric cues through motion parallax and structural consistency across frames. Moreover, when extended to temporal settings, naively concatenating historical visual observations without re-anchoring each frame to the language instruction causes the instruction signal to be progressively diluted by the expanding visual token sequence. 
These limitations motivate our instruction-anchored streaming framework \ourMethod{}.

\subsection{Instruction-Anchored Temporal Modeling}
\label{sec:temporal_modeling}

A naive approach to temporal aggregation is to concatenate historical visual observations into a single growing sequence, which introduces two compounding problems.
First, the sequence length grows linearly with the temporal horizon, making both training memory consumption and inference latency prohibitive for real-time control.
Second, as visual tokens accumulate, the language instruction maybe be progressively overshadowed, causing the model to lose track of the task goal over long horizons.
To address these issues, we propose instruction-anchored temporal modeling in a streaming fashion, shown in Figure~\ref{fig:method}.
Specifically, each visual observation is paired with the task instruction, forming an instruction-anchored temporal unit.
We apply bidirectional attention within each instruction-observation pair to capture multimodal interactions, and causal attention across pairs to model temporal dependencies.

First, we define each time step's
input as an \textit{atomic temporal unit} that jointly encodes the
multi-view visual observations and the language instruction:
\begin{equation}
\mathbf{u}_t = (\mathbf{V}_t,\, {l}_t)
\end{equation}
where ${l}_t$ denotes the language instruction corresponding
to time $t$.
By treating $\mathbf{u}_t$ as an indivisible unit, the instruction
remains persistently anchored to its associated visual context at
each time step, preventing instruction forgetting regardless of
the temporal horizon length.
At the current time step, we define the streaming frames as the latest $T$ sampling observations, including the current one. The input sequence is then constructed by concatenating $T$ atomic temporal units:
\begin{equation}
\mathbf{U} = [\mathbf{u}_{t-T+1},\, \mathbf{u}_{t-T+2},\, \ldots,\, \mathbf{u}_t]
\end{equation}

To capture both cross-modal fusion within each time step and
temporal dependencies across time steps, we organize the attention
structure over $\mathbf{U}$ from two levels.

\textbf{Intra-pair Bidirectional Attention.}
Within each atomic unit $\mathbf{u}_\tau$, all tokens from
$\mathbf{V}_\tau$ and ${l}_\tau$ attend to each other
bidirectionally:
\begin{equation}
\mathbf{h}_\tau = \mathrm{Attn}_{\mathrm{bi}}\!\left(\mathbf{V}_\tau,\, {l}_\tau\right)
\end{equation}
This ensures thorough cross-modal fusion between the multi-view
visual observations and the language instruction at each individual
time step, producing a semantically grounded representation
$\mathbf{h}_\tau$ for downstream temporal reasoning.

\textbf{Inter-pair Causal Attention.}
Across atomic units, the fused representation $\mathbf{h}_\tau$
attends to all preceding units via causal attention:
\begin{equation}
\mathbf{o}_t = \mathrm{Attn}_{\mathrm{causal}}\!\left(\mathbf{h}_{t-T+1},\, \ldots,\, \mathbf{h}_t\right)
\end{equation}
This allows the model to aggregate temporal context from past
observations in an autoregressive manner, while the causal
structure ensures that future frames do not leak into past
representations during training.
The final output $\mathbf{o}_t$ is then used to condition action
generation of $\mathbf{a}_t \sim \pi_\theta(\cdot \mid \mathbf{o}_t)$. 

\paragraph{Parameter-free Temporal Extension.}
A key advantage of this design is that it introduces no additional parameters.
The hierarchical attention pattern is entirely realized by restructuring the attention mask over the existing VLA backbone.
The inter-pair causal attention over an extended token sequence is handled naturally by the LLM's length extrapolation capability, allowing {\ourMethod} to inherit all pre-trained weights without modification. Specifically, we only extend the input token sequence for multi-frame inputs and assign extended position embeddings to all frame tokens. We then use our inter-pair causal mask to enforce intra-frame self-attention within each individual frame and inter-frame causal attention across consecutive frames, restricting the input to follow the temporal order strictly.
This ensures that the rich vision-language representations learned during pre-training are effectively preserved.

\subsection{Random-Interval Streaming Training Strategy}
\label{sec:training}


In real-robot deployment, observations arrive asynchronously and at
variable frame rates. A model trained with fixed temporal intervals
becomes brittle to such variation, since the temporal statistics at
test time differ systematically from those encountered during training.
We address this mismatch with a dedicated training strategy that
improves robustness to temporal irregularity.


\paragraph{Random-interval Sampling.}
During training, rather than sampling streaming frames at a strictly
fixed interval, we introduce a random interval to
improve robustness. Concretely, given a base inter-frame interval
$\bar{\delta}$, we add a random perturbation $\epsilon \sim
\mathcal{U}(-\Delta, +\Delta)$ at each sampling step, yielding
a perturbed interval $\delta = \bar{\delta} + \epsilon$, which is
clipped to $[\delta_{\min}, \delta_{\max}]$.
The $T$ streaming frames are then sampled from the streaming buffer
at steps of $\delta$.
This exposes the model to a diverse distribution of temporal spacings
around the nominal interval, preventing it from over-relying on
fixed-interval temporal cues and improving generalization to the
asynchronous observation rates encountered during real-robot deployment. 
In the bottom-right corner of Figure~\ref{fig:method}, a schematic illustrates the proposed Random-Interval Sampling strategy, where historical frames are selected at randomized offsets $t{-}\delta_1$ and $t{-}\delta_1{-}\delta_2$ relative to the current frame $t$, rather than at fixed intervals.

\paragraph{Temporal Masking.}
To further align training with streaming inference, we adopt a
temporal masking strategy. Given a full sequence of $T$ streaming
frames, we randomly sample a masking count $k \in \{0, 1, \ldots,
T-1\}$. When $k = 0$, the complete sequence is visible to the model.
When $k > 0$, the earliest $k$ frames are masked, leaving only the
most recent $T - k$ frames accessible. Combined with causal attention
masking, this simulates the incremental observation pattern of
streaming inference.

\paragraph{Streaming Inference.}
With our streaming design, \ourMethod~enables strong temporal modeling
without introducing significant additional computational overhead.
Given a sequential stream of observations at timestamps
$\{t_0, t_1, \ldots, t_N\}$, at the initial timestamp $t_0$, the
model takes the current frame $\mathbf{u}_{t_0}$ as input, predicts
the action $\mathbf{a}_{t_0}$, and stores the resulting fused
representation $\mathbf{h}_{t_0}$ in the KV-Cache.
At each subsequent timestamp $t_n$ ($n > 0$), only the newly arriving
frame $\mathbf{u}_{t_n}$ needs to be encoded. Its representation
$\mathbf{h}_{t_n}$ then attends to the cached historical
representations $\{\mathbf{h}_{t_0}, \ldots, \mathbf{h}_{t_{n-1}}\}$
via cross-attention, eliminating redundant re-computation over past
frames.
After each step, the KV-Cache is updated by appending
$\mathbf{h}_{t_n}$, which is directly reused for temporal modeling
at all future timestamps.
This design eliminates redundant recomputation of past frames and makes
\ourMethod~well-suited for long-horizon robot manipulation tasks.


\section{Experiments}

To comprehensively evaluate \ourMethod, we organize experiments
around four core questions:
(1) Can it handle both memory-dependent and precise perception-dependent
manipulation tasks on real robots? (Sec.~\ref{sec:exp_real})
(2) How does \ourMethod~compare with the advanced single-frame VLA methods
on the LIBERO simulation benchmark? (Sec.~\ref{sec:exp_libero})
(3) How does each design choice contribute to overall performance?
(Sec.~\ref{sec:ablation})
(4) How robust is \ourMethod~under varying temporal sampling
intervals and cross frames?
(Sec.~\ref{sec:ablation})

\subsection{Experimental Details}

\noindent\textbf{ Benchmark.}
We evaluate \ourMethod~on the LIBERO benchmark~\cite{libero},
which comprises four task suites of increasing complexity:
LIBERO-Spatial, LIBERO-Object, LIBERO-Goal, and LIBERO-Long,
each containing 10 tasks with 50 trials
per task. We also evaluate {\ourMethod} on the CALVIN benchmark~\cite{calvin} to demonstrate its temporal modeling capacity.
We follow the standard evaluation protocol and report success
rates averaged as the evaluation metric. Besides, we provide details of the real-robot setup for our real-world experiments in the technical appendices.

\noindent\textbf{Training \& Inference.}
\ourMethod~is built on a
pre-trained Vision-Language-Action model $\pi_{0.5}$~\cite{pi05}, and keep the same settings including learning rate and optimizer.
We fully fine-tune the pre-trained weights and introduce no additional
parameters and temporal modeling is achieved through
the attention mask. Specifically, 
we adopt the same optimizer and learning
rate schedule as $\pi_{0.5}$.
The number of streaming frames is set to $T{=}3/5$ during training.
For random-interval streaming training, the inter-frame
interval $\delta$ is sampled uniformly from
$[\delta_{\min}, \delta_{\max}] = [3, 7]$ at each training step.
All experiments are conducted on 8 NVIDIA H100 GPUs
with a batch size of 256 for LIBERO benchmark with 30k iterations and with a batch size of 128 for all real-robot tasks with 50k iterations.
At inference time, \ourMethod~operates in a streaming manner,
maintaining a rolling buffer of the $T$ most recent
observation-instruction pairs with the interval of $\delta$.
The inter-frame interval is fixed at $\delta{=}5$ for simulation
and sampled uniformly from $\delta \sim \mathcal{U}[3, 7]$
for real-robot deployment.

\begin{figure*}[t!]
\centering
\includegraphics[width=0.99\linewidth]{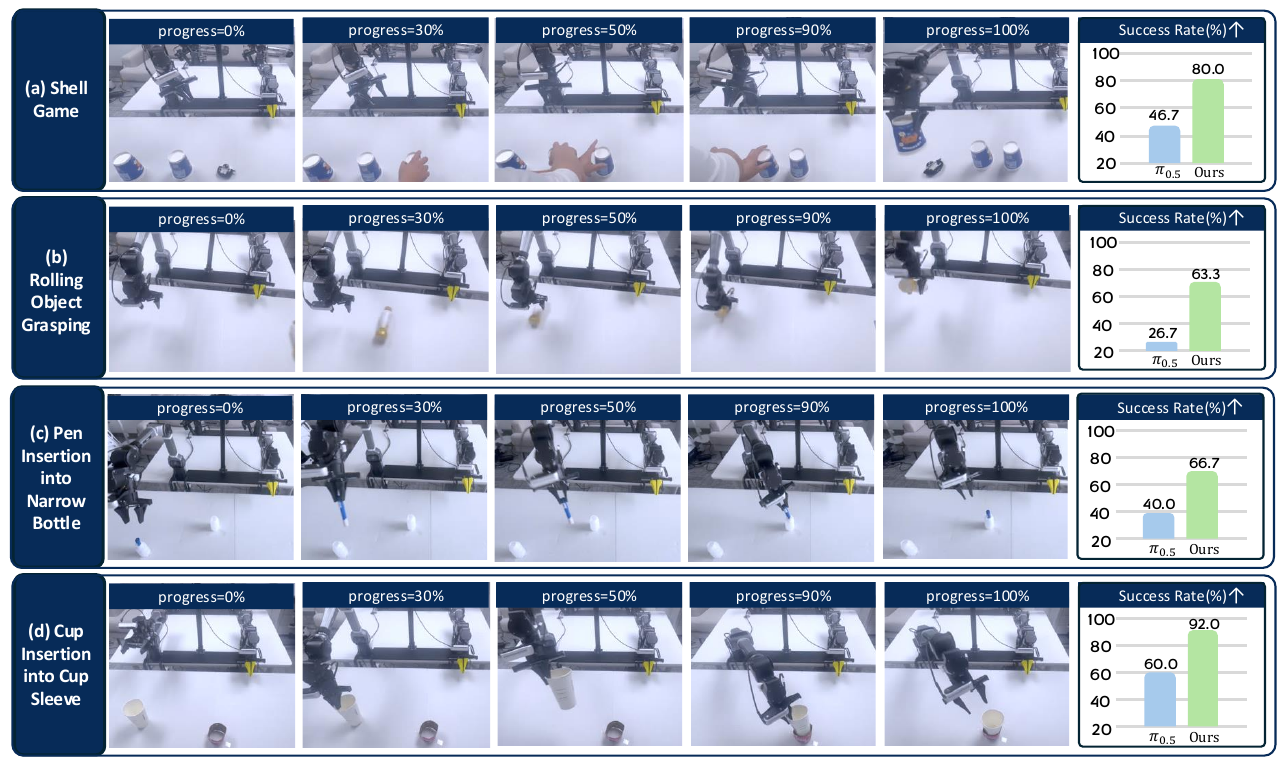}
\vspace{-5pt}
\caption{Visualization of real-world tasks and comparison of real-world performance. \textbf{Left}: Visualization of memory-dependent and precise perception-dependent real-world tasks. \textbf{Right}: Performance comparison of real-world tasks between ${\pi_{0.5}}$ and {\ourMethod}.
}
\label{fig:exp}
\vspace{-10pt}
\end{figure*}

\subsection{Real-Robot Manipulation}
\label{sec:exp_real}

Complex real-robot manipulation usually requires both precise geometric
perception and persistent temporal awareness
across interaction steps. To verify the effectiveness of
\ourMethod~in addressing these demands, we conduct real-robot
experiments spanning two complementary task categories: 1)~\textit{Memory-Dependent Tasks},
which require the robot to retain and act upon contextual cues
accumulated over long horizons;
2)~\textit{Precise Perception-Dependent Tasks}, which focus on the model's
ability to perceive fine-grained geometric relationships from
multi-frame observations. Each category consists of two tasks, evaluated  with success rate (\%).  

\noindent\textbf{Memory-Dependent Tasks.}
These tasks require the robot to recall information from earlier
observations to complete a later step, a capability fundamentally
beyond the reach of single-frame inference.
As shown in Figure~\ref{fig:exp}(a) and (b), we evaluate on two specific tasks.
\textit{i) Shell Game}: following a shell-game
protocol, the robot first observes an object being concealed
beneath one of several cups, after which the cups are shuffled.
The robot then retrieves the correct cup by recalling which
one concealed the object, a task that requires cross-frame memory.
{\ourMethod} achieves substantial improvements on both tasks
(+\textbf{36.6}\% on Rolling Object Grasping and
+\textbf{33.3}\% on Cup Hiding and Retrieval),
demonstrating that instruction-anchored temporal modeling
effectively preserves task-relevant memory across extended
horizons. \textit{ii) Rolling Object Grasping}: the robot need to track a
continuously moving object across frames and grasp it at the
appropriate moment. Since the object's position shifts over time,
a single frame provides no motion context and thus cannot support
reliable interception.
For more details about the tasks, please refer to our appendix.

\noindent\textbf{Precise Perception-Dependent Tasks.}
These tasks require accurate 3D perception from single-frame observations, where a single frame
provides insufficient geometric context for reliable execution.
We evaluate on two specific tasks.
\textit{i) Pen Insertion into Narrow Bottle}: the robot need to
guide a pen into a narrow bottle opening, demanding sub-centimeter
spatial precision that a single frame cannot reliably provide due
to the absence of multi-view depth cues.
\textit{ii) Cup Insertion into Cup Sleeve}: the robot need to
precisely align and insert a cup into a designated cup sleeve,
where accurate depth estimation and pose alignment across frames
are essential to avoid misplacement.
As shown in Figure~\ref{fig:exp}(c) and (d), \ourMethod~consistently
outperforms the single-frame $\pi_{0.5}$ baseline across both tasks 
+\textbf{26.7}\% on Pen Insertion into Narrow Bottle) and (+\textbf{32.0}\% on Cup Insertion into Cup Sleeve, confirming that temporal aggregation provides
critical geometric cues unavailable from any single observation.
For more details about the tasks, please refer to our appendix.

\subsection{LIBERO Simulation Benchmark}
\label{sec:exp_libero}

Beyond real-robot experiments, we further validate \ourMethod~on
the LIBERO benchmark~\cite{libero}, a widely adopted
simulation suite comprising four task suites: LIBERO-Spatial, LIBERO-Object, LIBERO-Goal, and
LIBERO-Long. This enables a systematic comparison against
existing VLA methods under standardized evaluation conditions.
We place particular emphasis on \textbf{LIBERO-Long}, which
requires executing sequences of 2 or more sub-tasks and most
directly tests long-horizon temporal memory, aligning closely
with the memory-dependent challenges addressed by \ourMethod.
As shown in Table~\ref{tab:libero}, the LIBERO benchmark is largely
saturated, with the baseline already exceeding 95\% success rate on
most suites, making further gains increasingly difficult to obtain.
Nevertheless, \ourMethod~achieves an average improvement of 1.4\%
over the single-frame $\pi_{0.5}$ baseline across all four suites,
demonstrating the effectiveness of temporal reasoning even in
near-saturated regimes. The gain is most pronounced on LIBERO-Long
(+\textbf{2.6}\%), where the lack of temporal memory is particularly
detrimental.
Notably, although a single frame already provides sufficient context for many LIBERO-Goal tasks, \ourMethod~still achieves a promising gain of 2.8\% on LIBERO-Goal, suggesting that the benefits of multi-frame modeling extend beyond memory-dependent tasks to perception-heavy scenarios involving dynamic object interactions and subtle state transitions. We observe no improvement on LIBERO-Spatial, where task success is often determined by static spatial relations that are already well captured in a single frame. In such cases, additional temporal context may introduce intermediate motion cues that are less relevant to the final geometric configuration. Overall, these results confirm that the temporal reasoning capability introduced by \ourMethod~generalizes robustly from real-robot manipulation to simulation environments.

\begin{table}[t]
    \centering
    \caption{Performance comparison on LIBERO~\citep{libero}.
    Success rates (\%) are reported across four suites. }
\setlength{\tabcolsep}{8pt}
\resizebox{1.0\linewidth}{!}{
    \begin{tabular}{c|cccc|c}
        \toprule
        Method & Libero-Spatial & Libero-Object & Libero-Goal & Libero-Long & Avg. Success Rate \\
        \midrule
        Diffusion Policy~\cite{chi2025diffusion} & 78.3 & 92.5 & 68.3 & 50.5 & 72.4 \\
        Octo~\cite{octo} & 78.9 & 85.7 & 84.6 & 51.1 & 75.1 \\
        SpatialVLA~\cite{spatialvla} & 88.2 & 89.9 & 78.6 & 55.5  & 71.7 \\
        TraceVLA~\cite{zheng2024tracevla} & 84.6 & 85.2 & 75.1 & 54.1  & 74.8 \\
        OpenVLA~\cite{openvla} & 84.7 & 88.4 & 79.2 & 53.7  & 75.9 \\
        CoT-VLA~\cite{zhao2025cot} & 87.5 & 91.6 & 87.6 & 69.0  & 81.1 \\
        $\pi_0$-FAST*~\cite{pertsch2025fast} & 96.4 & 96.8 & 88.6 & 60.2 & 85.0 \\
        SmolVLA~\cite{smolvla} & 93.0 & 94.0 & 91.0 & 77.0 & 88.8 \\
        GR00T-N1~\cite{gr00t} & 94.4 & 97.6 & 93.0 & 90.6 & 93.9 \\
        UniVLA~\cite{univla} & 95.4 & 98.8 & 93.6 & 94.0 & 95.4 \\
        FLOWER~\cite{reuss2025flower} & 97.1 & 96.7 & 95.6 & 93.5 & 95.7 \\
        CronusVLA~\cite{li2026towards} & 90.1 & 94.7 & 91.3 & 68.7 & 86.2 \\
        TriVLA~\cite{liu2025trivla} & 91.2 & 93.8 & 89.8 & 73.2 & 87.0 \\
        4D-VLA~\cite{zhang20254d} & 93.8 & 92.8 & 95.6 & 86.5 & 92.2 \\
        CogACT~\cite{li2024cogact} & 87.5 & 90.2 & 80.2 & 53.2 & 77.8 \\
        ST-$\pi$~\cite{ma2026st} & 98.4 & 98.3 & 96.9 & 94.3 & 97.3 \\
        MemoryVLA~\cite{shi2025memoryvla} & 98.4 & 98.4 & 96.4 & 93.4 & 96.5\\
        $\pi_0$~\cite{pi0} & 96.8 & 98.8 & 95.8 & 85.2 & 94.2 \\
        $\pi_{0.5}$~\cite{pi05} & 98.8 & 98.2 & 96.8 & 92.4 & 96.9 \\
        \midrule
        \rowcolor{blue!4}
        \ourMethod~($T=3$)  & \textbf{98.6} & 98.6 & 98.6 & 93.8 & 97.5 \\
        \rowcolor{blue!8}
        \ourMethod~($T=5$) & \textbf{98.8} & \textbf{99.8} & \textbf{99.6} & \textbf{95.0} & \textbf{98.3}  \\
        \bottomrule
    \end{tabular}}
    \label{tab:libero}
    \vspace{-10pt}
\end{table}

\subsection{CALVIN Benchmark}
\label{sec:exp_calvin}

To further evaluate the long-horizon temporal reasoning capabilities of {\ourMethod}, we conduct experiments on the CALVIN benchmark~\cite{calvin}, a challenging multi-task robotic manipulation suite that requires executing sequences of up to 5 consecutive tasks. Unlike single-task benchmarks, CALVIN explicitly tests the ability to compose actions across extended temporal horizons and maintain task progress over lengthy interactions.

As shown in Table~\ref{tab:calvin}, \ourMethod~($T=5$) achieves an average sequence length of {4.547}, substantially outperforming both the single-frame $\pi_{0.5}$ baseline (4.313) and MemoryVLA~\cite{shi2025memoryvla} (4.090). While all methods perform comparably on the first task, {\ourMethod} maintains significantly higher success rates at later stages ({85.0}\% vs. 79.5\% for $\pi_{0.5}$ and 69.4\% for MemoryVLA on the 5th task). Notably, MemoryVLA, despite being designed for temporal memory, exhibits severe performance degradation in later sequence positions, suggesting that its memory mechanism struggles with error accumulation over long horizons. In contrast, {\ourMethod} demonstrates robust and consistent gains at every sequence position, indicating the effectiveness of multi-frame temporal modeling.

\begin{table}[t]
    \centering
    \caption{Performance comparison on CALVIN~\citep{calvin}. 
    Success rates (\%) at each sequence position and average sequence length are reported.}
\setlength{\tabcolsep}{18pt}
\resizebox{1.0\linewidth}{!}{
    \begin{tabular}{c|ccccc|c}
        \toprule
        Method & 1 & 2 & 3 & 4 & 5 & Avg. Seq. Len \\
        \midrule
        MemoryVLA~\cite{shi2025memoryvla} & 94.8 & 87.4 & 81.4 & 75.9 & 69.4 & 4.090 \\
        $\pi_{0.5}$~\cite{pi05} & 94.2 & 88.7 & 85.7 & 83.2 & 79.5 & 4.313 \\
        \rowcolor{blue!8}
        \ourMethod~($T=5$) & \textbf{96.9} & \textbf{93.6} & \textbf{90.7} & \textbf{88.5} & \textbf{85.0} & \textbf{4.547}  \\
        \bottomrule
    \end{tabular}}
    \label{tab:calvin}
    \vspace{-10pt}
\end{table}

\subsection{Ablation Studies}
\label{sec:ablation}

Unless otherwise specified, all ablation studies are conducted
on the LIBERO benchmark to validate the key design choices of
{\ourMethod}. 

\noindent\textbf{Effectiveness of Instruction-Anchored Temporal Modeling.}
We ablate the instruction-anchored temporal modeling at two levels: intra-pair
(image-text fusion within each temporal unit) and inter-pair
(aggregation across temporal units).
As shown in Table~\ref{tab:ablation_attn}, both dimensions
contribute meaningfully to the final performance.
For intra-pair attention, replacing bidirectional with causal
attention leads to a consistent performance drop across all
four LIBERO suites at every $T$, with the gap widening as $T$
increases ($-$\textbf{5.6}\% on LIBERO-Long at $T{=}5$).
Causal intra-pair attention may prevent visual tokens from
attending to the instruction, breaking the semantic coupling
that \ourMethod~relies upon and causing instruction forgetting
over long horizons.
For inter-pair attention, its benefit is evident when comparing
$T{=}1$ (no inter-pair attention) against $T{=}5$ under
bidirectional intra-pair attention, yielding a
\textbf{+1.8}\% average gain and \textbf{+3.0}\% on
LIBERO-Long. This illustrates causal cross-temporal aggregation effectively
accumulates task-relevant context as more streaming frames
are processed.
These results verify that both bidirectional
intra-pair fusion and causal inter-pair aggregation are
essential components of \ourMethod's instruction-anchored
temporal modeling.

\noindent\textbf{Effect of Random-Interval Streaming Training.}
We study the effect of random-interval streaming training in the LIBERO simulation benchmark by comparing models trained with a fixed interval $\delta{=}1$ and those trained with random temporal intervals.
As shown in Table~\ref{tab:ablation_interval}, random-interval training consistently improves performance over fixed-interval training under the same temporal length.
For $T{=}3$, the average success rate increases from 96.4 to 97.5, while for $T{=}5$, it further improves from 97.0 to 98.3.
The gains are especially clear on long-horizon tasks, suggesting that exposure to diverse temporal spacings helps the policy better capture temporal dependencies.
Moreover, the interval $\delta$ also controls the frequency of streaming inference: $\delta{=}1$ requires the policy to process every incoming frame, whereas a larger interval, e.g., $\delta{=}5$, invokes inference only once every five frames.
Therefore, maintaining strong performance under random and large intervals is important for reducing online inference overhead and leaving more time for action execution in streaming robot control.

\noindent\textbf{Cross-Stream Generalization.}
We investigate whether a model trained with $T{=}5$ historical
frames generalizes to inference with fewer frames ($T{=}3$
and $T{=}1$).
As reported in Table~\ref{tab:ablation_stream}, the
trained model with $T{=}5$ retains strong performance when evaluated
at $T{=}3$, with only a marginal degradation, suggesting that
the model has internalized temporal patterns that remain
partially effective even under reduced context.
Performance at $T{=}1$ degrades more noticeably, yet still
surpasses the $\pi_{0.5}$ single-frame baseline, indicating that
temporal structure learned during training provides a residual
benefit even when no historical frames are available at
inference time. 

\begin{table}[t!]
\centering
\caption{Ablation on attention direction. {Intra-pair}
denotes attention between image and text tokens within each
temporal unit. {Inter-pair} denotes attention across
temporal units and is always causal. Results are average success
rates (\%) on the LIBERO benchmark.}
\label{tab:ablation_attn}
\setlength{\tabcolsep}{8pt}
\resizebox{1.0\linewidth}{!}{
\begin{tabular}{cllccccc}
\toprule
{ $T$~(Streaming Frames)} & {Intra-Pair} & {Inter-Pair} & {Spatial} & {Object} & {Goal} & {Long} & { Avg. Success Rate} \\
\midrule
\multirow{2}{*}{$1$}
  & Causal        & ---           & 97.4 & 97.8 & 96.8 & 91.0 & 95.8 \\
  & Bidirectional & ---           & 98.4 & 98.2 & 97.4 & 92.0 & 96.5 \\
\midrule
\multirow{2}{*}{$3$}
  & Causal        & Causal        & 98.6 & 96.0 & 95.4 & 91.0 & 95.3 \\
  & Bidirectional & Causal        & 98.8 & 98.6 & 98.6 & 93.8 & 97.5 \\
\midrule
\multirow{2}{*}{$5$}
  & Causal        & Causal        & 98.0 & 97.6 & 96.8 & 90.6 & 95.5 \\
  & Bidirectional & Causal        & \textbf{98.8} & \textbf{99.8} & \textbf{99.6} & \textbf{95.0} & \textbf{98.3} \\
\bottomrule
\end{tabular}}
\vspace{-10pt}
\end{table}

\begin{table}[t!]
\centering
\caption{Effect of random-interval streaming training. Models are
trained with fixed ($\delta{=}1$) or random interval
($\delta \sim \mathcal{U}[3, 7]$) and evaluated at
$\delta \in \{1, 3, 5\}$. 
}
\label{tab:ablation_interval}
\setlength{\tabcolsep}{12pt}
\resizebox{1.0\linewidth}{!}{
\begin{tabular}{lcccccccc}
\toprule
{Training $\delta$} & {Training $T$} &  {Test $T$} & {Spatial} & {Object} & {Goal} & {Long} & {Avg. Success Rate} \\
\midrule
$\pi_{0.5}$ ($\delta{=}1$) & 1 & 1 & 98.4 & 98.2 & 97.4 & 92.0 & 96.5   \\
\midrule
\multirow{2}{*}{Fixed ($\delta{=}1$)}
 &3 & 3 & 98.0 & 98.4 & 96.8 & 92.2 & 96.4 \\
 & 5 & 5 & 98.4 & 97.4 & 98.6 & 93.4 & 97.0  \\
\midrule
\multirow{2}{*}{Random $\delta $}
& 3  & 3 & 98.8 & 98.6 & 98.6 & 93.8 & 97.5  \\
&5  & 5 &  \textbf{98.8} & \textbf{99.8} & \textbf{99.6} & \textbf{95.0} & \textbf{98.3} \\
\bottomrule
\end{tabular}}
\vspace{-10pt}
\end{table}

\begin{table}[t!]
\centering
\caption{Cross-stream generalization. A model trained with $T{=}5$
is evaluated at $T \in \{1, 3, 5\}$. Results are average success
rates (\%) on LIBERO-Long.}
\label{tab:ablation_stream}
\setlength{\tabcolsep}{15pt}
\resizebox{1.0\linewidth}{!}{
\begin{tabular}{lcccccc}
\toprule
{Training $T$} & {Test $T$} & {Spatial} & {Object} & {Goal} & {Long} & {Avg. Success Rate} \\
\midrule
$\pi_{0.5}$ (baseline)  & 1 & 98.4 & 98.2 & 97.4 & 92.0 & 96.5  \\
\midrule
\multirow{2}{*}{$T{=}3$}
  & 1 & 98.0 & 98.8 & 97.8 & 93.4 & 97.0 \\
  & 3 & 98.8 & 98.6 & 98.6 & 93.8 & 97.5 \\
\midrule
\multirow{3}{*}{$T{=}5$}
  & 1 & 97.4 & 98.8 & 98.4 & 93.6 & 97.1 \\
  & 3 & \textbf{99.0} & 98.8 & 98.0 & 93.8 & 97.4 \\
  & 5  & \textbf{98.8} & \textbf{99.8} & \textbf{99.6} & \textbf{95.0} & \textbf{98.3} \\
\bottomrule
\end{tabular}}
\vspace{-10pt}
\end{table}

\section{Conclusion}
In this work, we present {\ourMethod}, a streaming multi-modal temporal modeling framework that equips VLA models with robust temporal awareness for robot manipulation. 
By treating \textit{(image, text)} pairs as atomic temporal units and combining intra-pair bidirectional attention with inter-pair causal attention, {\ourMethod} effectively captures cross-frame geometric context and maintains persistent instruction grounding over long horizons, without introducing additional parameters. 
To reduce the mismatch between fixed-interval training and asynchronous deployment, we further introduce Random-Interval Streaming Training, which exposes the model to diverse temporal spacings during training. 
Extensive experiments on real-robot manipulation tasks and the LIBERO benchmark show that {\ourMethod} consistently outperforms the single-frame baseline, with notable gains on perception-sensitive and memory-dependent tasks. 
We hope {\ourMethod} provides a simple and effective recipe for endowing future VLA models with persistent temporal reasoning.

{\small
\bibliographystyle{plain}
\bibliography{ref}

@String(CVPR  = {IEEE Conf. Comput. Vis. Pattern Recog.})

@String(ECCV  = {Eur. Conf. Comput. Vis.})

@String(NeurIPS = {Adv. Neural Inform. Process. Syst.})

@String(ICLR  = {Int. Conf. Learn. Represent.})

@String(AAAI  = {AAAI})

@String(CVPR  = {CVPR})

@String(ECCV  = {ECCV})

@String(NeurIPS = {NeurIPS})

@String(ICLR  = {ICLR})

@article{dynamicvla,
  title   = {DynamicVLA: A Vision-Language-Action Model for Dynamic Object Manipulation},
  author  = {Xie, Haozhe and Wen, Beichen and Zheng, Jiarui and Chen, Zhaoxi and Hong, Fangzhou and Diao, Haiwen and Liu, Ziwei},
  journal = {arXiv preprint arXiv:2601.22153},
  year    = {2026}
}

@article{smolvla,
  title   = {Smolvla: A vision-language-action model for affordable and efficient robotics},
  author  = {Shukor, Mustafa and Aubakirova, Dana and Capuano, Francesco and Kooijmans, Pepijn and Palma, Steven and Zouitine, Adil and Aractingi, Michel and Pascal, Caroline and Russi, Martino and Marafioti, Andres and others},
  journal = {arXiv preprint arXiv:2506.01844},
  year    = {2025}
}

@inproceedings{pi0,
  title     = {$\pi_0$: A Vision-Language-Action Flow Model for General Robot Control},
  author    = {Black, Kevin and Brown, Noah and Driess, Danny and Esmail, Adnan and Equi, Michael and Finn, Chelsea and Fusai, Niccolo and Groom, Lachy and Hausman, Karol and Ichter, Brian and others},
  booktitle = {RSS},
  year      = {2025}
}

@article{pi05,
  title   = {$\pi_{0.5}$: a Vision-Language-Action Model with Open-World Generalization},
  author  = {Intelligence, Physical and Black, Kevin and Brown, Noah and Darpinian, James and Dhabalia, Karan and Driess, Danny and Esmail, Adnan and Equi, Michael and Finn, Chelsea and Fusai, Niccolo and others},
  journal = {arXiv preprint arXiv:2504.16054},
  year    = {2025}
}

@inproceedings{openvla,
  title     = {{OpenVLA}: An Open-Source Vision-Language-Action Model},
  author    = {Kim, Moo Jin and Pertsch, Karl and Karamcheti, Siddharth and Xiao, Ted and Balakrishna, Ashwin and Nair, Suraj and Rafailov, Rafael and Foster, Ethan and Lam, Grace and Sanketi, Pannag and others},
  booktitle = {CoRL},
  year      = {2024}
}

@inproceedings{rt2,
  title     = {Rt-2: Vision-language-action models transfer web knowledge to robotic control},
  author    = {Zitkovich, Brianna and Yu, Tianhe and Xu, Sichun and Xu, Peng and Xiao, Ted and Xia, Fei and Wu, Jialin and Wohlhart, Paul and Welker, Stefan and Wahid, Ayzaan and others},
  booktitle = {CoRL},
  pages     = {2165--2183},
  year      = {2023}
}

@inproceedings{octo,
  title     = {{Octo}: An open-source generalist robot policy},
  author    = {Ghosh, Dibya and Walke, Homer and Pertsch, Karl and Black, Kevin and Mees, Oier and Dasari, Sudeep and Hejna, Joey and Kreiman, Tobias and Xu, Charles and others},
  booktitle = {RSS},
  year      = {2024}
}

@article{gr00t,
  author  = {Bjorck, Johan and Casta{\~n}eda, Fernando and Cherniadev, Nikita and Da, Xingye and Ding, Runyu and Fan, Linxi and Fang, Yu and Fox, Dieter and Hu, Fengyuan and Huang, Spencer and others},
  title   = {{GR00T} {N1:} An Open Foundation Model for Generalist Humanoid Robots},
  journal = {arXiv preprint arXiv:2503.14734},
  year    = {2025}
}

@article{lingbotvla,
  title   = {A Pragmatic VLA Foundation Model},
  author  = {Wu, Wei and Lu, Fan and Wang, Yunnan and Yang, Shuai and Liu, Shi and Wang, Fangjing and Zhu, Qian and Sun, He and Wang, Yong and Ma, Shuailei and others},
  journal = {arXiv preprint arXiv:2601.18692},
  year    = {2026}
}

@inproceedings{vlaadapter,
  title     = {Vla-adapter: An effective paradigm for tiny-scale vision-language-action model},
  author    = {Wang, Yihao and Ding, Pengxiang and Li, Lingxiao and Cui, Can and Ge, Zirui and Tong, Xinyang and Song, Wenxuan and Zhao, Han and Zhao, Wei and Hou, Pengxu and others},
  booktitle = AAAI,
  year      = {2025}
}

@article{ni2025swiftvla,
  title   = {SwiftVLA: Unlocking Spatiotemporal Dynamics for Lightweight VLA Models at Minimal Overhead},
  author  = {Ni, Chaojun and Chen, Cheng and Wang, Xiaofeng and Zhu, Zheng and Zheng, Wenzhao and Wang, Boyuan and Chen, Tianrun and Zhao, Guosheng and Li, Haoyun and Dong, Zhehao and others},
  journal = {arXiv preprint arXiv:2512.00903},
  year    = {2025}
}

@article{chen2025nanovla,
  title   = {NanoVLA: Routing Decoupled Vision-Language Understanding for Nano-sized Generalist Robotic Policies},
  author  = {Chen, Jiahong and Wang, Jing and Chen, Long and Cai, Chuwei and Lu, Jinghui},
  journal = {arXiv preprint arXiv:2510.25122},
  year    = {2025}
}

@inproceedings{reuss2025flower,
  title     = {Flower: Democratizing generalist robot policies with efficient vision-language-action flow policies},
  author    = {Reuss, Moritz and Zhou, Hongyi and R{\"u}hle, Marcel and Ya{\u{g}}murlu, {\"O}mer Erdin{\c{c}} and Otto, Fabian and Lioutikov, Rudolf},
  booktitle = {CoRL},
  year      = {2025}
}

@inproceedings{duan2025real,
  title     = {Real-time Iteration Scheme for Diffusion Policy},
  author    = {Duan, Yufei and Yin, Hang and Kragic, Danica},
  booktitle = {IROS},
  year      = {2025}
}

@article{pertsch2025fast,
  title   = {Fast: Efficient action tokenization for vision-language-action models},
  author  = {Pertsch, Karl and Stachowicz, Kyle and Ichter, Brian and Driess, Danny and Nair, Suraj and Vuong, Quan and Mees, Oier and Finn, Chelsea and Levine, Sergey},
  journal = {arXiv preprint arXiv:2501.09747},
  year    = {2025}
}

@inproceedings{univla,
  title     = {{UniVLA}: Learning to Act Anywhere with Task-centric Latent Actions},
  author    = {Bu, Qingwen and Yang, Yanting and Cai, Jisong and Gao, Shenyuan and Ren, Guanghui and Yao, Maoqing and Luo, Ping and Li, Hongyang},
  booktitle = RSS,
  year      = {2025}
}

@inproceedings{xvla,
  title     = {{X-VLA}: Soft-Prompted Transformer as Scalable Cross-Embodiment Vision-Language-Action Model},
  author    = {Zheng, Jinliang and Li, Jianxiong and Wang, Zhihao and Liu, Dongxiu and Kang, Xirui and Feng, Yuchun and Zheng, Yinan and Zou, Jiayin and Chen, Yilun and Zeng, Jia and others},
  booktitle = ICLR,
  year      = {2026}
}

@article{tinyvla,
  title   = {{TinyVLA}: Towards fast, data-efficient vision-language-action models for robotic manipulation},
  author  = {Wen, Junjie and Zhu, Yichen and Li, Jinming and Zhu, Minjie and Tang, Zhibin and Wu, Kun and Xu, Zhiyuan and Liu, Ning and Cheng, Ran and Shen, Chaomin and others},
  journal = {RAL},
  year    = {2025}
}

@inproceedings{act,
  title     = {Learning fine-grained bimanual manipulation with low-cost hardware},
  author    = {Zhao, Tony Z and Kumar, Vikash and Levine, Sergey and Finn, Chelsea},
  booktitle = {RSS},
  year      = {2023}
}

@inproceedings{libero,
  author    = {Liu, Bo and Zhu, Yifeng and Gao, Chongkai and Feng, Yihao and Liu, Qiang and Zhu, Yuke and Stone, Peter},
  booktitle = NeurIPS,
  pages     = {44776--44791},
  title     = {LIBERO: Benchmarking Knowledge Transfer for Lifelong Robot Learning},
  year      = {2023}
}

@article{calvin,
  author  = {Oier Mees and Lukas Hermann and Erick Rosete-Beas and Wolfram Burgard},
  title   = {{CALVIN}: A Benchmark for Language-Conditioned Policy Learning for Long-Horizon Robot Manipulation Tasks},
  journal = {RAL},
  year    = {2022}
}

@article{bevdet4d,
  title={Bevdet4d: Exploit temporal cues in multi-camera 3d object detection},
  author={Huang, Junjie and Huang, Guan},
  journal={arXiv preprint arXiv:2203.17054},
  year={2022}
}

@article{bevformer,
  title={Bevformer: learning bird’s-eye-view representation from lidar-camera via spatiotemporal transformers},
  author={Li, Zhiqi and Wang, Wenhai and Li, Hongyang and Xie, Enze and Sima, Chonghao and Lu, Tong and Yu, Qiao and Dai, Jifeng},
  journal=TPAMI,
  year={2024}
}

@inproceedings{wang2023exploring,
  title={Exploring object-centric temporal modeling for efficient multi-view 3d object detection},
  author={Wang, Shihao and Liu, Yingfei and Wang, Tiancai and Li, Ying and Zhang, Xiangyu},
  booktitle=CVPR,
  pages={3621--3631},
  year={2023}
}

@inproceedings{open,
  title={Open: Object-wise position embedding for multi-view 3d object detection},
  author={Hou, Jinghua and Wang, Tong and Ye, Xiaoqing and Liu, Zhe and Gong, Shi and Tan, Xiao and Ding, Errui and Wang, Jingdong and Bai, Xiang},
  booktitle=ECCV,
  year={2024},
}

@inproceedings{vggt,
  title={Vggt: Visual geometry grounded transformer},
  author={Wang, Jianyuan and Chen, Minghao and Karaev, Nikita and Vedaldi, Andrea and Rupprecht, Christian and Novotny, David},
  booktitle=CVPR,
  pages={5294--5306},
  year={2025}
}

@article{zhuo2025streaming,
  title={Streaming 4d visual geometry transformer},
  author={Zhuo, Dong and Zheng, Wenzhao and Guo, Jiahe and Wu, Yuqi and Zhou, Jie and Lu, Jiwen},
  journal={arXiv preprint arXiv:2507.11539},
  year={2025}
}

@article{chi2025diffusion,
  title={Diffusion policy: Visuomotor policy learning via action diffusion},
  author={Chi, Cheng and Xu, Zhenjia and Feng, Siyuan and Cousineau, Eric and Du, Yilun and Burchfiel, Benjamin and Tedrake, Russ and Song, Shuran},
  journal={The International Journal of Robotics Research},
  volume={44},
  number={10-11},
  pages={1684--1704},
  year={2025},
}

@inproceedings{li2026towards,
  title={Towards Efficient and Robust Manipulation via Multi-Frame Vision-Language-Action Modeling},
  author={Li, Hao and Yang, Shuai and Chen, Yilun and Chen, Xinyi and Yang, Xiaoda and Tian, Yang and Wang, Hanqing and Wang, Tai and Lin, Dahua and Zhao, Feng and others},
  booktitle=AAAI,
  year={2026}
}

@article{fung2025embodied,
  title={Embodied ai agents: Modeling the world},
  author={Fung, Pascale and Bachrach, Yoram and Celikyilmaz, Asli and Chaudhuri, Kamalika and Chen, Delong and Chung, Willy and Dupoux, Emmanuel and Gong, Hongyu and J{\'e}gou, Herv{\'e} and Lazaric, Alessandro and others},
  journal={arXiv preprint arXiv:2506.22355},
  year={2025}
}

@article{li2025comprehensive,
  title={A comprehensive survey on world models for embodied ai},
  author={Li, Xinqing and He, Xin and Zhang, Le and Wu, Min and Li, Xiaoli and Liu, Yun},
  journal={arXiv preprint arXiv:2510.16732},
  year={2025}
}

@article{xiao2023efficient,
  title={Efficient streaming language models with attention sinks},
  author={Xiao, Guangxuan and Tian, Yuandong and Chen, Beidi and Han, Song and Lewis, Mike},
  journal={arXiv preprint arXiv:2309.17453},
  year={2023}
}

@article{gong2026ace,
  title={Ace-brain-0: Spatial intelligence as a shared scaffold for universal embodiments},
  author={Gong, Ziyang and Luo, Zehang and Tang, Anke and Liu, Zhe and Fu, Shi and Hou, Zhi and Yang, Ganlin and Wang, Weiyun and Wang, Xiaofeng and Liu, Jianbo and others},
  journal={arXiv preprint arXiv:2603.03198},
  year={2026}
}

@article{chen2023longlora,
  title={Longlora: Efficient fine-tuning of long-context large language models},
  author={Chen, Yukang and Qian, Shengju and Tang, Haotian and Lai, Xin and Liu, Zhijian and Han, Song and Jia, Jiaya},
  journal={arXiv preprint arXiv:2309.12307},
  year={2023}
}

@article{shi2026streamingvla,
  title={StreamingVLA: Streaming Vision-Language-Action Model with Action Flow Matching and Adaptive Early Observation},
  author={Shi, Yiran and Guo, Dongqi and Zhao, Tianchen and Gao, Feng and Shi, Liangzhi and Yu, Chao and Mo, ZhiJian and Xiao, Qihua and Peng, XiaoShuai and Liao, Qingmin and others},
  journal={arXiv preprint arXiv:2603.28565},
  year={2026}
}

@article{shi2025memoryvla,
  title={Memoryvla: Perceptual-cognitive memory in vision-language-action models for robotic manipulation},
  author={Shi, Hao and Xie, Bin and Liu, Yingfei and Sun, Lin and Liu, Fengrong and Wang, Tiancai and Zhou, Erjin and Fan, Haoqiang and Zhang, Xiangyu and Huang, Gao},
  journal={arXiv preprint arXiv:2508.19236},
  year={2025}
}

@article{lu2026faster,
  title={FASTER: Rethinking Real-Time Flow VLAs},
  author={Lu, Yuxiang and Liu, Zhe and Fan, Xianzhe and Yang, Zhenya and Hou, Jinghua and Li, Junyi and Ding, Kaixin and Zhao, Hengshuang},
  journal={arXiv preprint arXiv:2603.19199},
  year={2026}
}

@article{spatialvla,
  title={Spatialvla: Exploring spatial representations for visual-language-action model},
  author={Qu, Delin and Song, Haoming and Chen, Qizhi and Yao, Yuanqi and Ye, Xinyi and Ding, Yan and Wang, Zhigang and Gu, JiaYuan and Zhao, Bin and Wang, Dong and others},
  journal={arXiv preprint arXiv:2501.15830},
  year={2025}
}

@article{wang2026real,
  title={Real-Time Robot Execution with Masked Action Chunking},
  author={Wang, Haoxuan and Zhang, Gengyu and Yan, Yan and Shang, Yuzhang and Kompella, Ramana Rao and Liu, Gaowen},
  journal={arXiv preprint arXiv:2601.20130},
  year={2026}
}

@inproceedings{zheng2024tracevla,
  title={Tracevla: Visual trace prompting enhances spatial-temporal awareness for generalist robotic policies},
  author={Zheng, Ruijie and Liang, Yongyuan and Huang, Shuaiyi and Gao, Jianfeng and Daum{\'e} III, Hal and Kolobov, Andrey and Huang, Furong and Yang, Jianwei},
  booktitle=ICLR,
  year={2024}
}

@article{jang2025contextvla,
  title={ContextVLA: Vision-Language-Action Model with Amortized Multi-Frame Context},
  author={Jang, Huiwon and Yu, Sihyun and Kwon, Heeseung and Jeon, Hojin and Seo, Younggyo and Shin, Jinwoo},
  journal={arXiv preprint arXiv:2510.04246},
  year={2025}
}

@article{li2026remem,
  title={ReMem-VLA: Empowering Vision-Language-Action Model with Memory via Dual-Level Recurrent Queries},
  author={Li, Hang and Shen, Fengyi and Chen, Dong and Yang, Liudi and Wang, Xudong and Shi, Jinkui and Bing, Zhenshan and Liu, Ziyuan and Knoll, Alois},
  journal={arXiv preprint arXiv:2603.12942},
  year={2026}
}

@article{cen2025rynnvla,
  title={Rynnvla-002: A unified vision-language-action and world model},
  author={Cen, Jun and Huang, Siteng and Yuan, Yuqian and Li, Kehan and Yuan, Hangjie and Yu, Chaohui and Jiang, Yuming and Guo, Jiayan and Li, Xin and Luo, Hao and others},
  journal={arXiv preprint arXiv:2511.17502},
  year={2025}
}

@article{li2024cogact,
  title={Cogact: A foundational vision-language-action model for synergizing cognition and action in robotic manipulation},
  author={Li, Qixiu and Liang, Yaobo and Wang, Zeyu and Luo, Lin and Chen, Xi and Liao, Mozheng and Wei, Fangyun and Deng, Yu and Xu, Sicheng and Zhang, Yizhong and others},
  journal={arXiv preprint arXiv:2411.19650},
  year={2024}
}

@article{ma2026st,
  title={ST-$\pi$: Structured SpatioTemporal VLA for Robotic Manipulation},
  author={Ma, Chuanhao and Zhou, Hanyu and Peng, Shihan and Li, Yan and Gu, Tao and Yan, Luxin},
  journal={arXiv preprint arXiv:2604.17880},
  year={2026}
}

@article{zhang20254d,
  title={4d-vla: Spatiotemporal vision-language-action pretraining with cross-scene calibration},
  author={Zhang, Jiahui and Chen, Yurui and Xu, Yueming and Huang, Ze and Zhou, Yanpeng and Yuan, Yu-Jie and Cai, Xinyue and Huang, Guowei and Quan, Xingyue and Xu, Hang and others},
  journal={arXiv preprint arXiv:2506.22242},
  year={2025}
}

@article{liu2025trivla,
  title={TriVLA: A Triple-System-Based Unified Vision-Language-Action Model with Episodic World Modeling for General Robot Control},
  author={Liu, Zhenyang and Gu, Yongchong and Zheng, Sixiao and Fu, Yanwei and Xue, Xiangyang and Jiang, Yu-Gang},
  journal={arXiv preprint arXiv:2507.01424},
  year={2025}
}

@inproceedings{zhao2025cot,
  title={Cot-vla: Visual chain-of-thought reasoning for vision-language-action models},
  author={Zhao, Qingqing and Lu, Yao and Kim, Moo Jin and Fu, Zipeng and Zhang, Zhuoyang and Wu, Yecheng and Li, Zhaoshuo and Ma, Qianli and Han, Song and Finn, Chelsea and others},
  booktitle=CVPR,
  year={2025}
}

@article{team2026ace,
  title={ACE-Brain-0.5: A Unified Embodied Foundational Model for Physical Agentic AI},
  author={Team, Brain and Gong, Ziyang and Gu, Haoming and Luo, Zehang and Zhang, Tianyi and Tao, Tao and Chi, Yixiao and Liu, Zhe and Zhu, Lingsi and Liu, Jingyuan and others},
  journal={arXiv preprint arXiv:2607.04426},
  year={2026}
}

@article{zhang2025dreamvla,
  title={Dreamvla: a vision-language-action model dreamed with comprehensive world knowledge},
  author={Zhang, Wenyao and Liu, Hongsi and Qi, Zekun and Wang, Yunnan and Yu, Xinqiang and Zhang, Jiazhao and Dong, Runpei and He, Jiawei and Lu, Fan and Wang, He and others},
  journal={arXiv preprint arXiv:2507.04447},
  year={2025}
}

@article{gao2026dreamdojo,
  title={DreamDojo: A Generalist Robot World Model from Large-Scale Human Videos},
  author={Gao, Shenyuan and Liang, William and Zheng, Kaiyuan and Malik, Ayaan and Ye, Seonghyeon and Yu, Sihyun and Tseng, Wei-Cheng and Dong, Yuzhu and Mo, Kaichun and Lin, Chen-Hsuan and others},
  journal={arXiv preprint arXiv:2602.06949},
  year={2026}
}

@inproceedings{liu2026drivepi,
  title={Drivepi: Spatial-aware 4d mllm for unified autonomous driving understanding, perception, prediction and planning},
  author={Liu, Zhe and Huang, Runhui and Yang, Rui and Yan, Siming and Wang, Zining and Hou, Lu and Lin, Di and Bai, Xiang and Zhao, Hengshuang},
  booktitle={Proceedings of the IEEE/CVF Conference on Computer Vision and Pattern Recognition},
  pages={3688--3698},
  year={2026}
}

@article{yuan2026fast,
  title={Fast-WAM: Do World Action Models Need Test-time Future Imagination?},
  author={Yuan, Tianyuan and Dong, Zibin and Liu, Yicheng and Zhao, Hang},
  journal={arXiv preprint arXiv:2603.16666},
  year={2026}
}

@article{song2026fast,
  title={Fast-dVLA: Accelerating Discrete Diffusion VLA to Real-Time Performance},
  author={Song, Wenxuan and Chen, Jiayi and Chen, Shuai and Wang, Jingbo and Ding, Pengxiang and Zhao, Han and Qin, Yikai and Zheng, Xinhu and Wang, Donglin and Wang, Yan and others},
  journal={arXiv preprint arXiv:2603.25661},
  year={2026}
}

@article{torne2026mem,
  title={Mem: Multi-scale embodied memory for vision language action models},
  author={Torne, Marcel and Pertsch, Karl and Walke, Homer and Vedder, Kyle and Nair, Suraj and Ichter, Brian and Ren, Allen Z and Wang, Haohuan and Tang, Jiaming and Stachowicz, Kyle and others},
  journal={arXiv preprint arXiv:2603.03596},
  year={2026}
}

@article{bi2025motus,
  title={Motus: A unified latent action world model},
  author={Bi, Hongzhe and Tan, Hengkai and Xie, Shenghao and Wang, Zeyuan and Huang, Shuhe and Liu, Haitian and Zhao, Ruowen and Feng, Yao and Xiang, Chendong and Rong, Yinze and others},
  journal={arXiv preprint arXiv:2512.13030},
  year={2025}
}
}

\newpage

\appendix

\section{Technical Appendices}

The appendix provides supplementary details covering implementation
specifics of streaming inference, additional real-world experimental
results, qualitative visualizations, limitations, future directions,
and broader impacts. A video demonstration is included in the
supplemental materials.

\section{Implementation Details}

\begin{algorithm}[h!]
\caption{Streaming Inference of \ourMethod~with KV-Cache}
\label{algo:infer}
\begin{algorithmic}[1]
\STATE \textbf{Input:} observation stream $\{\mathbf{V}_{t_0}, \mathbf{V}_{t_1}, \ldots, \mathbf{V}_{t_N}\}$,
       language instruction ${l}$, maximum cache size $T$
\STATE \textbf{Output:} predicted action chunks
       $\mathcal{A}$

\STATE Initialize KV-Cache $\mathcal{C} \leftarrow \emptyset$,\
       cache length $L_c \leftarrow 0$,\ step $n \leftarrow 0$

\WHILE{$n \leq N$}
    \STATE Receive current observation $\mathbf{V}_{t_n}$
    \STATE Encode atomic unit $(\mathbf{V}_{t_n},\, {l})$
           with cached $\mathcal{C}$ to obtain fused representation
           $\mathbf{h}_{t_n}$
    \IF{$L_c + 1 > T$}
        \STATE Flush cache: $\mathcal{C} \leftarrow \emptyset$, $L_c \leftarrow 0$
    \ENDIF
    \STATE Update cache $\mathcal{C}$ and $L_c \leftarrow L_c + 1$
    \STATE Predict action $\mathcal{A}_n \sim
           \pi_\theta(\cdot \mid \mathbf{h}_{t_n},\, \mathcal{C})$
    \STATE Dispatch $\mathcal{A}_n$ to client 
    \STATE $n \leftarrow n + 1$
\ENDWHILE
\end{algorithmic}
\end{algorithm}

\subsection{Streaming Inference}

The pseudo-code for the streaming inference of \ourMethod~is provided
in \textbf{Algorithm 1}.
To evaluate inference efficiency, we conduct 20 trials on a real-robot
platform equipped with a single NVIDIA GeForce RTX~4090 GPU and report
the mean latency with standard deviation in Table~\ref{tab:time}.
The single-frame baseline incurs a latency of $94.4 \pm 3.4$\,ms.
Extending the temporal context to 3 frames adds only 3.5\,ms of
overhead ($97.9 \pm 5.1$\,ms total), and scaling further to 5, 8, and 10 frames
yields $103.6 \pm 6.3$\,ms, $110.9 \pm 10.2$\,ms, and $117.9 \pm 16.5$\,ms, respectively. The 5-frame setting only brings an increase of merely 9.2\,ms over the
baseline.
These results confirm that \ourMethod~achieves efficient streaming
inference, maintaining high responsiveness even as the temporal
context grows.

\begin{table*}[h!]
\caption{The inference time of different streaming frames in streaming inference.}
\vspace{-5pt}
\small
\setlength{\tabcolsep}{22pt}
\centering
\begin{tabular}{c|c}
\toprule
Streaming Frames & Inference Time (ms) \\
\midrule
1 & $94.4 \pm 3.4$ \\
3 & $97.9 \pm 5.1$ \\
5 & $103.6 \pm 6.3$ \\
8 & $110.9 \pm 10.2$ \\
10 & $117.9 \pm 16.5$ \\
\bottomrule
\end{tabular}
\label{tab:time}
\vspace{-10pt}
\end{table*}

\subsection{Real-world Experiments}

\begin{figure}[h!]
    \centering
    \includegraphics[width=0.99\linewidth]{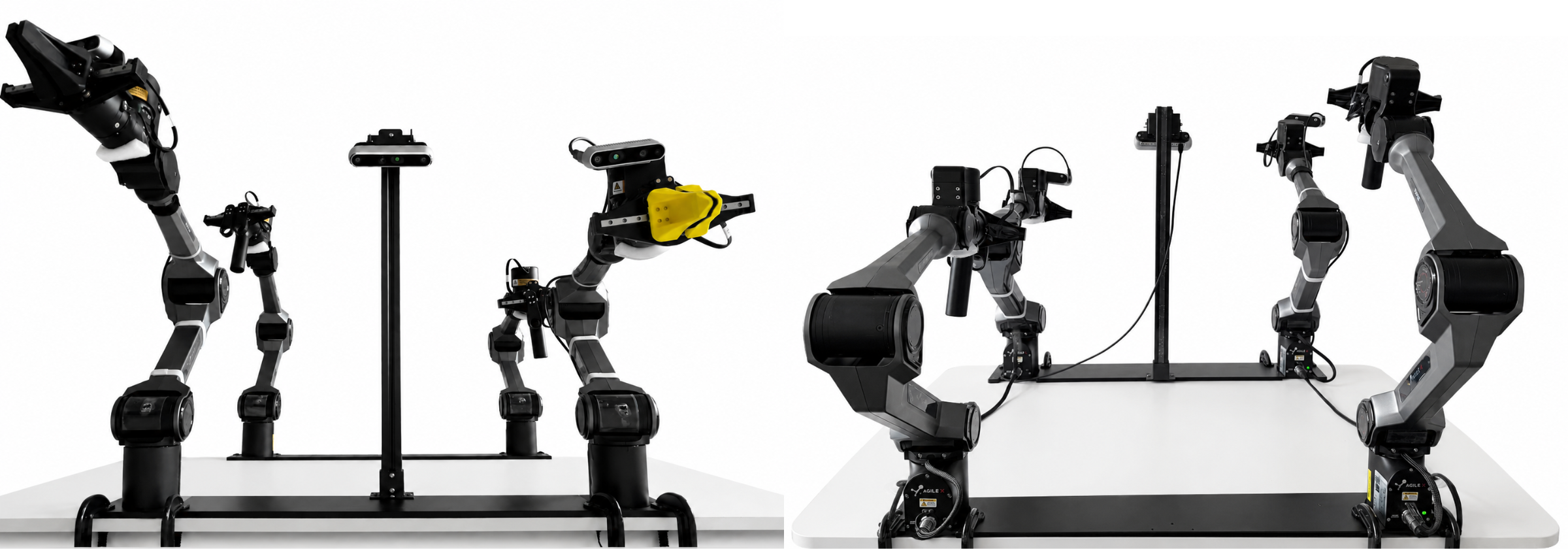}
    \caption{AgileX PiperX robotic arms.}
    \label{fig:arm}
      \vspace{-10pt}
\end{figure}

\paragraph{Hardware Setup.}
We use the AgileX PiperX 6-DoF robotic arms shown in Figure~\ref{fig:arm}.
The system follows the Aloha-style design~\cite{act}, with leader arms for human teleoperation
and follower arms for data collection and rollout. We use three cameras: one front-view camera (RealSense D455) and two wrist-mounted cameras (RealSense D435) on the follower arms.

\paragraph{Tasks.}
We evaluate four real-world tasks to demonstrate the superiority of {\ourMethod}. Specifically, we design two spatial-precision tasks (i.e., ``Cup Insertion into Cup Sleeve'', ``Pen Insertion into Narrow Bottle'') and memory-dependent tasks (i.e., ``Rolling Object Grasping'', ``Shell Game''). For each task, we collect 100 demonstration episodes using human teleportation at 30 FPS. 

The language instructions for the tasks are as follows:
\begin{itemize}
\item \textbf{Shell Game}: ``Pick up the cup that contains the hidden object after the shuffles.''
\item \textbf{Rolling Object Grasping}: ``Pick up the rolling bottle.''
\item \textbf{Pen Insertion into Narrow Bottle}: ``Insert the pen from one bottle into another bottle.''
\item \textbf{Cup Insertion into Cup Sleeve}: ``Pick up the paper cup and put it into the cup sleeve.''
\end{itemize}

\paragraph{Evaluation.}
To validate the real-world performance of {\ourMethod}, we design detailed comparative experiments for each task. The score is defined as 1 point for success and 0 points for failure. We adopt success rate as the evaluation metric.

For ``Shell Game'', we design 15 different patterns. Each pattern is shown in Table~\ref{tab:guess_pattern}. We provide detailed performance comparison of each trial in Table~\ref{tab:real_guess}. {\ourMethod} achieves 33.3\% improvement compared to $\pi_{0.5}$.

For ``Pick up the Rolling Bottle'', we evaluate performance with 30 trails. We provide detailed performance comparison of each trial in Table~\ref{tab:real_pick}. This task is a highly dynamic task and needs to have the ability to use temporal information to predict the motion of objects. {\ourMethod} outperforms $\pi_{0.5}$ by 36.6\%, demonstrating the effectiveness of proposed streaming temporal modeling.

For ``Pen Insertion into Narrow Bottle'', we set 3 different patterns. ``M-F'': move the pen from the middle cup to the top right cup. ``M-M'': move the pen from the left cup to the right cup. ``F-M'': move the pen from the left right cup to the middle cup. We evaluate performance with 10 trails on each pattern. We provide detailed performance comparison of each trial in Table~\ref{tab:real_pen}. This task not only requires high precise perception to insert the pen into a narrow bottle but also needs temporal information to determine which bottle to put the pen in. Benefited from introduced temporal modeling, {\ourMethod} is better able to handle this task than $\pi_{0.5}$.

\begin{table}[t!]
\caption{Performance comparison of each trial on ``Shell Game'' task. S.R. denotes success rate.}
\resizebox{1.0\linewidth}{!}{
  \centering
  \setlength{\tabcolsep}{10pt}
  \begin{tabular}{l|*{15}{c}|c}
    \toprule
    Model & 1 & 2 & 3 & 4 & 5 & 6 & 7 & 8 & 9 & 10 & 11 & 12 & 13 & 14 & 15 & S.R. (\%) \\
    \midrule
    $\pi_{0.5}$ 
    & 1 & 0 & 1 & 0 & 0 
    & 1 & 1 & 1 & 1 & 0 
    & 1 & 0 & 0 & 0 & 0 
    & 46.7\% \\
    {\ourMethod}
    & 1 & 1 & 1 & 0 & 1 
    & 1 & 1 & 1 & 1 & 1 
    & 1 & 0 & 1 & 1 & 0 
    & 80.0\% \\
    \bottomrule
  \end{tabular}
  }
\label{tab:real_guess}
  \vspace{-10pt}
\end{table}

\begin{table}[t!]
\caption{Performance comparison of each trial on ``Pick up the Rolling Bottle'' task. S.R. denotes success rate.}
\resizebox{1.0\linewidth}{!}{
  \centering
  \setlength{\tabcolsep}{3.5pt}
  \begin{tabular}{l|*{30}{c}|c}
    \toprule
    Model & 1 & 2 & 3 & 4 & 5 & 6 & 7 & 8 & 9 & 10 & 11 & 12 & 13 & 14 & 15 & 16 & 17 & 18 & 19 & 20 & 21 & 22 & 23 & 24 & 25 & 26 & 27 & 28 & 29 & 30 & S.R. (\%) \\
    \midrule
    $\pi_{0.5}$
    & 1 & 0 & 1 & 0 & 0 & 0 & 0 & 0 & 0 & 1 
    & 0 & 0 & 0 & 0 & 0 & 1 & 0 & 0 & 0 & 0 
    & 0 & 1 & 0 & 0 & 1 & 0 & 1 & 0 & 1 & 0 
    & 26.7 \\
    {\ourMethod}
    & 0 & 1 & 1 & 0 & 1 & 0 & 1 & 0 & 0 & 1 
    & 0 & 1 & 1 & 0 & 1 & 0 & 1 & 1 & 1 & 1 
    & 1 & 0 & 1 & 1 & 1 & 1 & 0 & 0 & 1 & 1 
    & 63.3 \\
    \bottomrule
  \end{tabular}
  }
\label{tab:real_pick}
  \vspace{-10pt}
\end{table}

\begin{table}[t!]
\caption{Performance comparison of each trial on ``Pen Insertion into Narrow Bottle'' task. S.R. denotes success rate.}
\small
\resizebox{1.0\linewidth}{!}{
  \centering
  \setlength{\tabcolsep}{4.5pt}
  \begin{tabular}{l|*{10}{c}|*{10}{c}|*{10}{c}|c}
    \toprule
    & \multicolumn{10}{c|}{M-F} & \multicolumn{10}{c|}{M-M} & \multicolumn{10}{c|}{F-M} & \multirow{2}{*}{S.R. (\%)} \\
    Model & 1 & 2 & 3 & 4 & 5 & 6 & 7 & 8 & 9 & 10 & 1 & 2 & 3 & 4 & 5 & 6 & 7 & 8 & 9 & 10 & 1 & 2 & 3 & 4 & 5 & 6 & 7 & 8 & 9 & 10 & \\
    \midrule
    $\pi_{0.5}$
    & 0 & 0 & 0 & 0 & 0 & 0 & 0 & 0 & 0 & 0 
    & 1 & 0 & 1 & 1 & 1 & 1 & 0 & 1 & 0 & 1 
    & 0 & 1 & 0 & 1 & 0 & 0 & 1 & 1 & 1 & 0 
    & 40.0 \\
    {\ourMethod}
    & 1 & 1 & 0 & 1 & 0 & 0 & 1 & 0 & 0 & 1 
    & 1 & 1 & 1 & 0 & 1 & 0 & 0 & 1 & 1 & 1 
    & 1 & 1 & 0 & 1 & 1 & 1 & 1 & 0 & 1 & 0 
    & 66.7 \\
    \bottomrule
  \end{tabular}
}
\label{tab:real_pen}
  \vspace{-10pt}
\end{table}

For ``Cup Insertion into Cup Sleeve'', we set 5 different positions of the cup sleeve: center, left-far, right-far, left-near, and right-near and evaluate performance with 5 trails on each position. We provide detailed performance comparison of each trial in Table~\ref{tab:real_cup}. {\ourMethod} achieves stronger perception capability compared to $\pi_{0.5}$, especially for distant locations of the cup sleeve.

\begin{table}[t!]
\caption{Performance comparison of each trial on “Cup Insertion into Cup Sleeve” task. S.R. denotes success rate.}
\small
\resizebox{1.0\linewidth}{!}{
  \centering
  \begin{tabular}{l|*{5}{*{5}{c}|}c}
    \toprule
    Model & \multicolumn{5}{c|}{Center} & \multicolumn{5}{c|}{Left-Far} & \multicolumn{5}{c|}{Right-Far} & \multicolumn{5}{c|}{Right-Near} & \multicolumn{5}{c|}{Left-Near} & S.R. (\%) \\
    \midrule
    $\pi_{0.5}$ & 
    1 & 0 & 1 & 1 & 1 & 
    0 & 0 & 0 & 1 & 0 & 
    0 & 0 & 0 & 0 & 0 & 
    1 & 1 & 1 & 1 & 1 & 
    1 & 1 & 1 & 1 & 1 & 
    60.0 \\
    \midrule
    {\ourMethod} & 
    1 & 1 & 1 & 1 & 1 & 
    1 & 1 & 1 & 0 & 1 & 
    1 & 1 & 0 & 1 & 1 & 
    1 & 1 & 1 & 1 & 1 & 
    1 & 1 & 1 & 1 & 1 & 
    92.0 \\
    \bottomrule
  \end{tabular}
  }
\label{tab:real_cup}
  \vspace{-10pt}
\end{table}

\begin{table}[t!]
\caption{Each pattern of the Shell Game. `L', `M', and `R' is the left, middle, and right position, respectively. A$\rightarrow$ B denotes that move the cup that contains the target object from A to B.}
\resizebox{1.0\linewidth}{!}{
  \centering
  \begin{tabular}{*{15}{c}}
    \toprule
     1 & 2 & 3 & 4 & 5 & 6 & 7 & 8 & 9 & 10 & 11 & 12 & 13 & 14 & 15 \\
    \midrule
    L$\rightarrow$M & L$\rightarrow$R & L$\rightarrow$M$\rightarrow$R & L$\rightarrow$M$\rightarrow$L & L$\rightarrow$R$\rightarrow$M & M$\rightarrow$L & M$\rightarrow$L$\rightarrow$M & M$\rightarrow$R & M$\rightarrow$R$\rightarrow$M & M$\rightarrow$L$\rightarrow$R & R$\rightarrow$M & R$\rightarrow$M$\rightarrow$R &
    R$\rightarrow$L & R$\rightarrow$L$\rightarrow$M & R$\rightarrow$L$\rightarrow$R \\
    \bottomrule
  \end{tabular}
  }
\label{tab:guess_pattern}
  \vspace{-10pt}
\end{table}

\begin{figure}[t!]
    \centering
    \includegraphics[width=0.99 \linewidth]{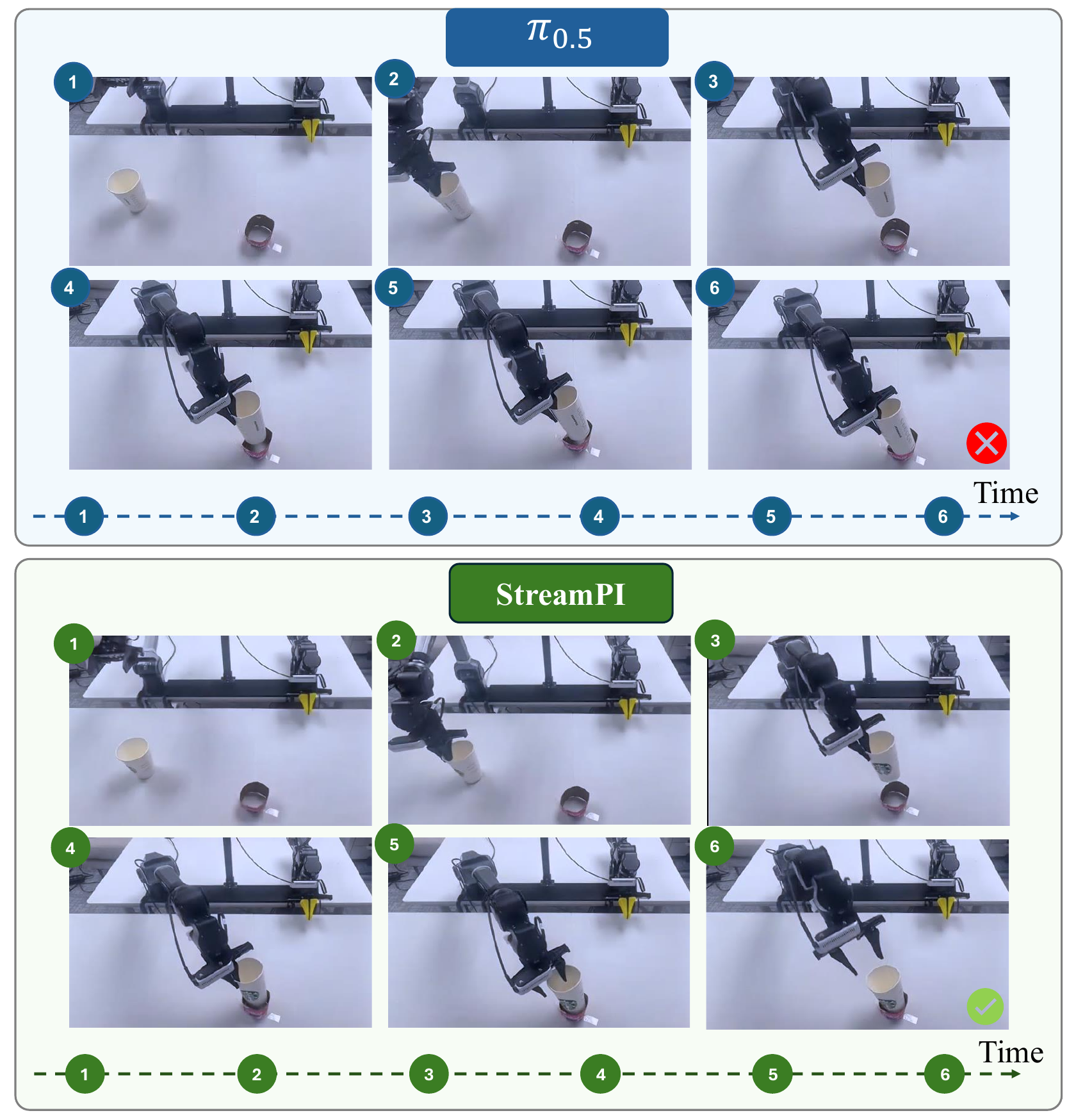}
    \caption{Comparison of ``Cup Insertion into Cup Sleeve''.}
    \label{fig:cmp_cup}
      \vspace{-15pt}
\end{figure}

\begin{figure}[t!]
    \centering
    \includegraphics[width=0.99 \linewidth]{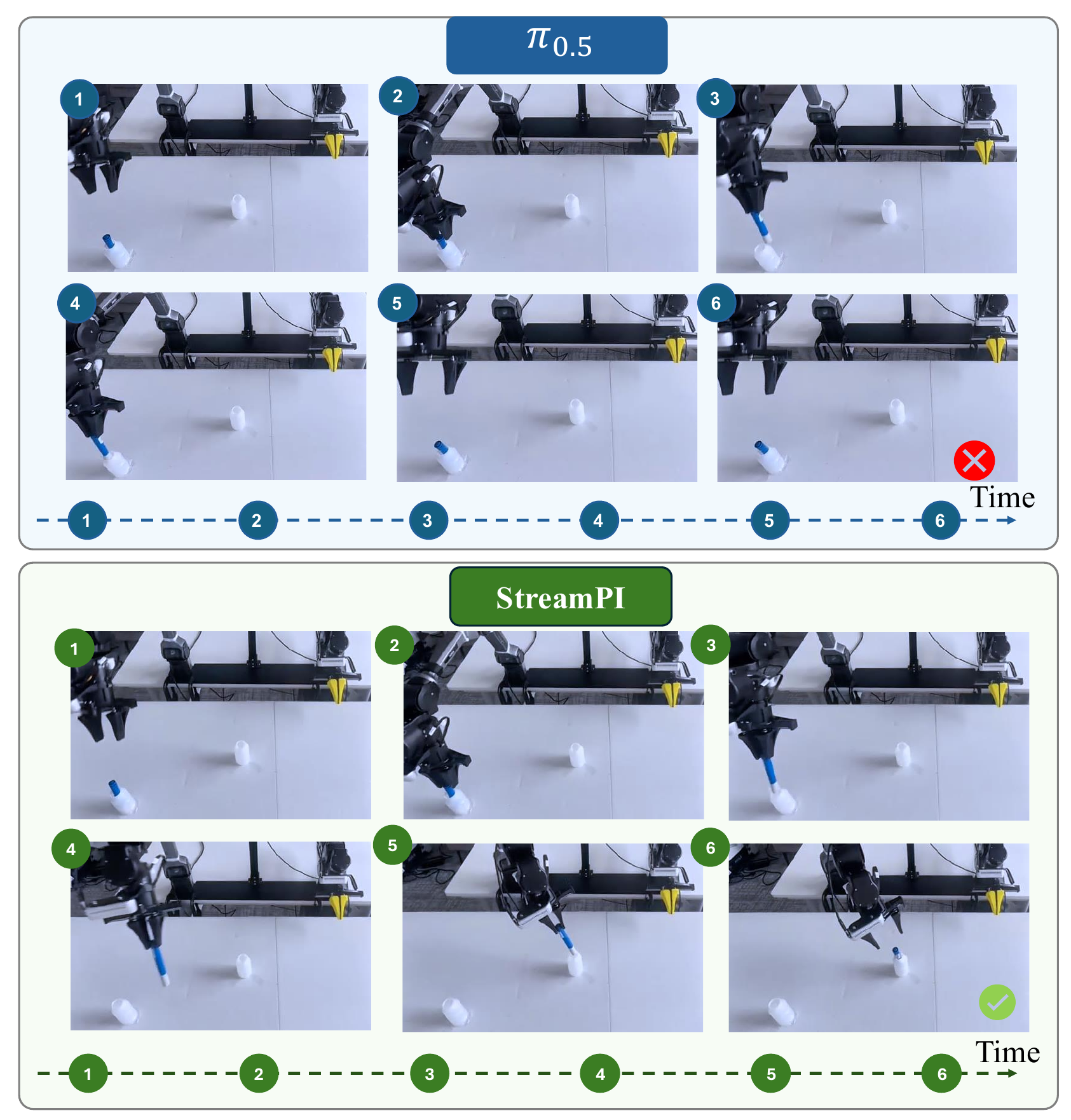}
    \caption{Comparison of ``Pen Insertion into Narrow Bottle''.}
    \label{fig:cmp_insert}
      \vspace{-15pt}
\end{figure}

\begin{figure}[t!]
    \centering
    \includegraphics[width=0.99 \linewidth]{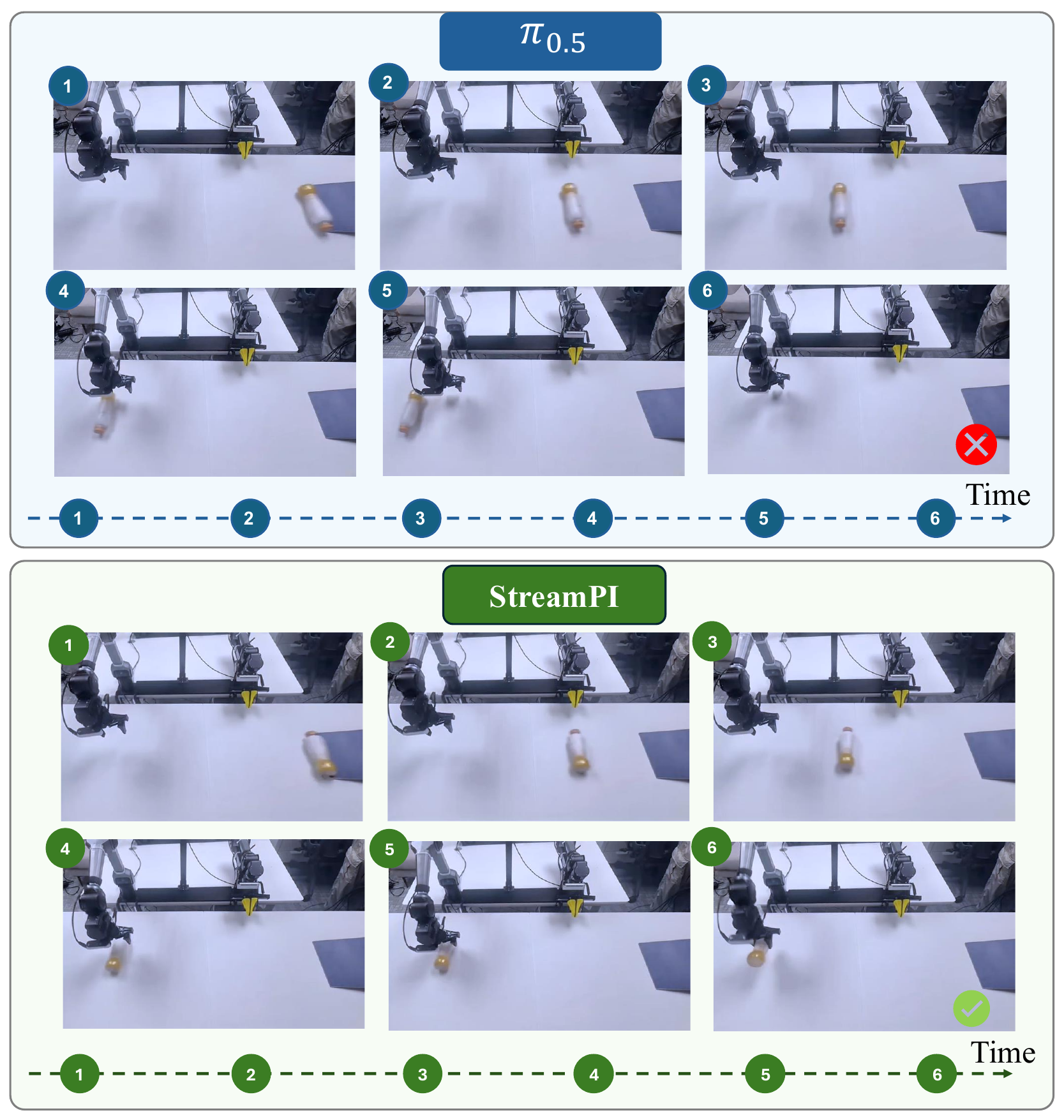}
    \caption{Comparison of ``Pick up the rolling bottle''.}
    \label{fig:cmp_pick}
    \vspace{-15pt}
\end{figure}

\section{Visualization Comparisons}
We provide visualization comparison of $\pi_{0.5}$ and {\ourMethod} on ``Cup Insertion into Cup Sleeve'', ``Pen Insertion into Narrow Bottle'', and ``Rolling Object Grasping'' tasks. 
\paragraph{Cup Insertion into Cup Sleeve.}
As shown in Figure~\ref{fig:cmp_cup}, $\pi_{0.5}$ struggles to accurately estimate the spatial position of distant objects, making it difficult to place the cup precisely in the cup sleeve. In contrast, {\ourMethod} accurately locate objects and place the cup successfully. 
\paragraph{Pen Insertion into Narrow Bottle.}
As shown in Figure~\ref{fig:cmp_insert}, ``Pen Insertion into Narrow Bottle'' needs both precise perception and temporal information. Therefore, $\pi_{0.5}$ cannot determine whether the current state is the beginning or the end of a task because it lacks temporal information, causing it to mistake picking up the pen for the moment it puts the pen into the bottle. {\ourMethod} successfully pick up the pen from one bottle and put it into another bottle, demonstrating the effectiveness of {\ourMethod} in temporal modeling. 
\paragraph{Pick up the rolling bottle.}
For ``Pick up the rolling bottle'', the slow reaction of $\pi_{0.5}$ results in missing the rolling bottle because it is unable to use temporal information to determine the bottle's motion properties. As shown in Figure~\ref{fig:cmp_pick}, Compared to $\pi_{0.5}$, {\ourMethod} can generate the corresponding action in advance by predicting the movement of objects, thus successfully catching the rolling bottle.

\begin{figure}[t!]
    \centering
    \includegraphics[width=0.99\linewidth]{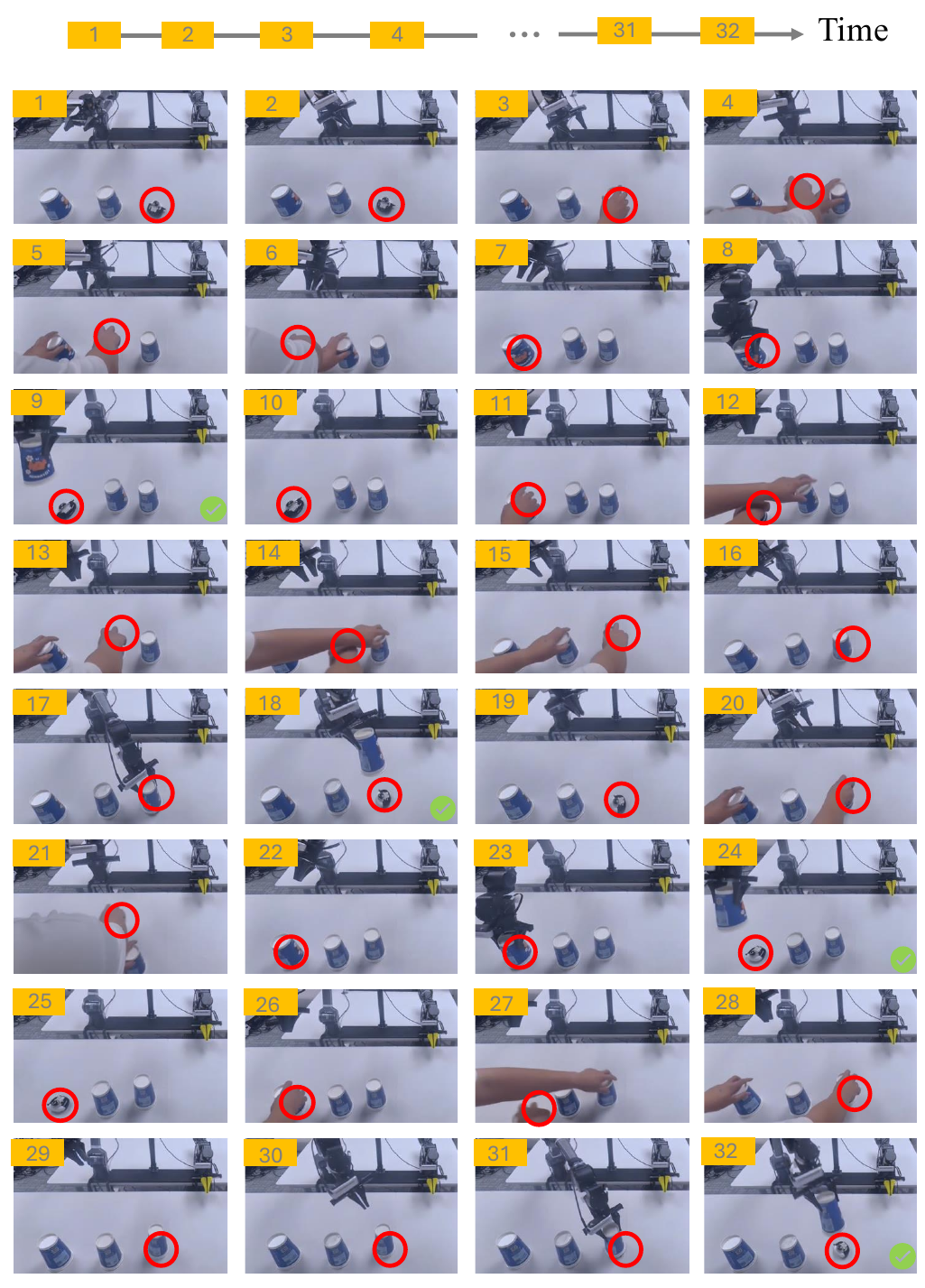}
    \caption{The continuous qualitative results of ``Shell Game''. For convenience, we use a red circle to indicate which cup contains the object. }
    \label{fig:vis_guess}
\end{figure}

\paragraph{Shell Game.}
To fully illustrate the effectiveness of {\ourMethod}, we provide a qualitative visualization of the ``Shell Game'' performed four times consecutively in Figure ~\ref{fig:vis_guess}. For convenience, we use a red circle to indicate which cup contains the object. Despite exchanging cups multiple times, {\ourMethod} can still accurately guess which cup contains the target object.

\section{Limitations and Future Work}
\noindent\textbf{Limitations.} Despite the effectiveness of {\ourMethod} in efficient temporal modeling for VLAs, it has several limitations. First, since training requires loading all frames, the training cost of {\ourMethod} in handling extremely long temporal horizons will be unacceptable. Second, the random-interval streaming training, while improving robustness to variable frame rates, does not fully address extreme asynchrony in real-robot deployment.

\noindent\textbf{Future Work.} In the future, we will design a more efficient training framework to support longer horizons~(>100 frames) with lower computational cost. Then, we plan to introduce adaptive KV cache pruning to maintain representation quality for ultra-long temporal horizons with negligible additional inference time.

\section{Broader Impacts}
{\ourMethod} advances robot manipulation by enabling efficient, robust temporal modeling without additional parameters, lowering the barrier for deploying strong foundation models in real-robot systems. Its streaming design and robustness to asynchronous observations make it suitable for real-world embodied tasks, improving efficiency and safety in human-robot interaction. However, the widespread deployment of more capable manipulation robots with our temporal modeling may affect labor markets in routine industrial tasks. Additionally, ensuring the robustness of VLAs with temporal modeling on the real-robot to out-of-distribution scenarios is critical to avoid safety risks in real-world operation, which we will address in future work.


\end{document}